\documentclass[fleqn,10pt]{wlscirep}
\usepackage[utf8]{inputenc}
\usepackage[T1]{fontenc}
\usepackage{comment}
\usepackage{acronym} 
\usepackage{cite}
\usepackage{subfig}
\usepackage{wrapfig}
\usepackage[binary-units=true]{siunitx}
\usepackage{textcomp}
\usepackage{flafter} 
\usepackage{lineno}

\usepackage{pdfpages}

\title{Finding the Gap: Neuromorphic Motion Vision in Cluttered Environments}

\author[1,2,3*]{Thorben Schoepe}
\author[4]{Ella Janotte}
\author[5]{Moritz B. Milde}
\author[6]{Olivier J.N. Bertrand}
\author[6]{Martin Egelhaaf}
\author[1,2,3]{Elisabetta Chicca}
\affil[1]{Faculty of Technology and Cognitive Interaction Technology Center of Excellence (CITEC), Bielefeld University, Germany.}
\affil[2]{Bio-Inspired Circuits and Systems (BICS) Lab. Zernike Institute for Advanced Materials (Zernike Inst Adv Mat), University of Groningen, Netherlands.}
\affil[3]{CogniGron (Groningen Cognitive Systems and Materials Center), University of Groningen, Netherlands.}
\affil[4]{Event Driven Perception for Robotics, Italian Institute of Technology, iCub facility, Genoa, Italy.}
\affil[5]{International Centre for Neuromorphic Systems, MARCS Institute, Western Sydney University, Penrith, Australia.}
\affil[6]{Neurobiology, Faculty of Biology, Bielefeld University, Bielefeld, Germany.}
\affil[*]{t.schoepe@rug.nl}

\begin{abstract}
Many animals meander in environments and avoid collisions. How the underlying neuronal machinery can yield robust behaviour in a variety of environments remains unclear. In the fly brain, motion-sensitive neurons indicate the presence of nearby objects and directional cues are integrated within an area known as the central complex.
Such neuronal machinery, in contrast with the traditional stream-based approach to signal processing, uses an event-based approach, with events occurring when changes are sensed by the animal. Contrary to classical von Neumann computing architectures, event-based neuromorphic hardware is designed to process information asynchronously and in a distributed manner. Inspired by the fly brain, we model, for the first time, a neuromorphic closed-loop system mimicking essential behaviours observed in flying insects, such as meandering in clutter and crossing of gaps, both of which are also highly relevant for autonomous vehicles. We implemented our system both in software and on neuromorphic hardware. While moving through an environment, our agent perceives changes in its surroundings and uses this information for collision avoidance. The agent's manoeuvres result from a closed action-perception loop implementing probabilistic decision-making processes. This loop-closure is thought to have driven the development of neural circuitry in biological agents since the Cambrian explosion. 
In the fundamental quest to understand neural computation in artificial agents, we come closer to understanding and modelling biological intelligence by closing the loop also in neuromorphic systems. As a closed-loop system, our system deepens our understanding of processing in neural networks and their computations in both biological and artificial systems. With these investigations, we aim to set the foundations for neuromorphic intelligence in the future, moving towards leveraging the full potential of neuromorphic systems.
\end{abstract}

\begin{document}
%%%%%%%%%%%%%%%% ACROS %%%%%%%%%%%%%%%%%%%

\acrodef{sEMD}[sEMD]{spiking Elementary Motion Detector}
\acrodef{OF}[OF]{optic flow}
\acrodef{WTA}[WTA]{winner take all}
\acrodef{SNN}[SNN]{spiking neural network}
\acrodef{TDE}[TDE]{Time Difference Encoder}
\acrodef{DVS}[DVS]{Dynamic Vision Sensor}
\acrodef{WTA}[WTA]{Winner-Take-All}
\acrodef{LIF}[LIF]{Leaky Integrate and Fire}
\acrodef{SPTC}[SPTC]{Spatio-Temporal Correlation}
\acrodef{POIS}[POIS]{Poisson Spike Generators}
\acrodef{INT}[INT]{Integrator}
\acrodef{ET}[ET]{Escape Turn}
\acrodef{GI}[GI]{Global Inhibition}
\acrodef{MOT}[MOT]{Motor}
\acrodef{OFI}[OFI]{Optic Flow Integrator}
\acrodef{BIBI}[BIBI]{Brain Interface and Body Integrator}
\acrodef{CLE}[CLE]{Closed Loop Engine}
\acrodef{ROS}[ROS]{Robot Operating System}
\acrodef{EBC}[EBC]{Event-Based Camera}
\acrodef{ISI}[ISI]{Inter-Spike-Interval}

%\includepdf[pages=1-2]{coverletter_collision_avoidance.pdf}

\flushbottom
\maketitle
\thispagestyle{empty}
%\begin{linenumbers}
\section{Introduction}
\label{sec:intro}

While navigating through the environment, our proprioception informs us about our posture, our eyes look for a familiar direction or goal, and our ears watch-out for dangers. The brain deals with multiple data-streams in a continuous and parallel manner. Autonomous vehicles requiring to safely manoeuvre in their environment also have to deal with such high-dimensional data-streams which are conventionally acquired and analysed at a fixed sampling frequency. A fixed sampling frequency limits the temporal resolution of data-processing and the amount of data which can be processed. To address these limitations, two approaches can be combined. First, data-streams can be sparsified by sending only information when an observed quantity changes, i.e. when it is required. Second, the data-stream can be processed in a parallel and asynchronous fashion.
This calls for an alternative approach to sensing and computing which, much like the brain, acquires and processes information completely asynchronously and in a distributed network of computing elements, e.g. neurons and synapses.
To fully demonstrate the advantages of this approach we use the example of autonomous navigation as it is well studied and algorithmically understood in a variety of environments be they water \cite{Kelasidi2019}, ground \cite{Floreano2014}, air \cite{Barca2013}, or space \cite{Rybus2018}. In the last decades, part of the engineering community has sought inspiration from animals \cite{Floreano2014, Pandey2017, Serres2018}. For example, flying insects such as bees and flies share the same requirements as light-weight flying vehicles manoeuvring in various habitats from almost object-free terrains \cite{Dickinson2014} to overly cluttered forests \cite{Baird2016} via human-made landscapes. They need to avoid collisions to prevent damaging their wings \cite{Mountcastle2016} and they accomplish this task by using limited neuronal resources (less than 1M \cite{Witthoft1967} and 100k \cite{Zheng2018} neurons for honeybees and fruit-flies respectively). At the core of this machinery is a well-described subset of neurons responding to the apparent motion of surrounding objects \cite{Borst2019, Fu2019computationalmodelscollision}. While the animal translates in its environment, the responses of such neurons provide estimates to the time-to-contact to nearby objects by approximating the apparent motion of the objects on the retina (i.e. the optic flow \cite{MartinEgelhaaf2012}). These neurons are thought to steer the animal away from obstacles \cite{Bertrand2015, Cope2016, Lecoeur2019TheRO, Mauss2020optic_flow} or toward gaps \cite{Baird2016gap, Ravi2019, Ravi2020} resulting in a collision-free path.
\\
\indent The collision avoidance machinery in insects is thought to be driven by a large array of motion-sensitive neurons, distributed in an omnidirectional visual field. These neurons operate asynchronously. Hence, biology has found an asynchronous and distributed solution to the problem of collision avoidance. We seek to emulate such a solution in bio-inspired neuromorphic hardware which has the advantage of being low-volume and low-power. More importantly, it also requires an asynchronous and parallel information processing implementation yielding a better understanding of neural computation.
\\
\indent To date, most of the mimics of the collision avoidance machinery rely on traditional cameras from which every pixel at every time point (i.e. at a fixed sampling frequency) needs to be processed \cite{Bertrand2015, Cope2016,Li2017,Serres2018, Zingg_etal10, Blosch_etal10}. The processing occurs even when nothing is changing in the agent's surroundings. This constant processing leads to a dense stream of data and consequently a high energy consumption. To reduce this, an efficient means of communication can be employed, such as action potentials observed in biological neural circuits. Action potentials or spikes enable to transmit information only when necessary, i.e. event-driven.
In an analogous way, event-based cameras send events asynchronously only when a change in luminance over time is observed \cite{Posch2010, Lichtsteiner2008, Brandli2014, Posch2014, Son2017}.
This sampling scheme is referred to as Lebesgue sampling \cite{Astrom_Bernhardsson02}.
Contrary to frame-based cameras, which employ Riemann sampling \cite{Astrom_Bernhardsson02}, bandwidth and power demands are significantly reduced (see Section~\nameref{sec:vision} for more details). \\
\indent
Open-loop collision avoidance based on optic-flow can use event-streams \cite{Benosman_etal13, Conradt2015, Milde2015, Liu2017, Rueckauer2016, Gallego2018, Haessig_etal18, Martel_etal15, Milde2018sEMD}(for more detailed comparison of mentioned approaches refer to \cite{Milde2018sEMD}) and an insect-inspired motion pathway has been suggested for collision avoidance \cite{Milde2018sEMD}.
Closed-loop collision avoidance behaviour have been demonstrated previously using fully conventional sensory-processing (frame-based sensor and CPUs/GPUs) approaches \cite{Zingg_etal10, Blosch_etal10} (for extensive review please refer to \cite{Serres_Ruffier17, Fu_etal19}). These insect-inspired approaches reduce the computational demands for collision avoidance by reducing the bandwidth of the visual input.
This reduction is achieved by collapsing the visual field into a left and right components. Later processing only needs to compare left versus right signals. These approaches, however, are hardwired processing of visual features. The hard-coded features may not be relevant in other environments. 
Mixed-system (event-based camera and conventional processing) approaches \cite{MullerConradt_eDVS2011, Gallego2018}, on the other hand, do not reduce the visual input by separating left-right signal pathways, but utilise event-based cameras which only transmit changes.
In contrast to biological systems, they do not, however, leverage the advantages of event-based processing until the actuation of the motors.
Finally, fully neuromorphic (event-based camera and parallel, asynchronous processing) approaches \cite{Milde2017, Kreiser_etal18} rely on spike-based information processing from sensing to actuation of motors. To date, these approaches rely on hardwired, deterministic decision making processing. The hard-coded decisions, i.e. creating a reflex-like machine, may lead to sub-optimal decisions when multiple directions to avoid collisions are viable.
% To avoid collisions one may plan its path in a known environment (e.g. by using maps) or derive collision-free path from sensory inputs, such as absolute distance information (e.g. LIDAR) or use apparent motion of objects on the retina, i.e. \acf{OF} which scales with relative distance. 
Here, we aim for the first time at closing the action-perception loop \cite{Serres_Ruffier17, Fu_etal19, Indiveri_Sandamirskaya19},
%Milde_etal21
while explicitly extracting insect-inspired visual features, making active decisions, and using neuromorphic spike-based computation from sensing to actuation. 
\indent Inspired by the collision avoidance algorithm proposed for flies and bees, we developed a \ac{SNN}\footnote{Spiking Neural Network: Massively parallel network consisting of populations of spike-based artificial neurons and synapses.} that profits from the parsimony of event-based cameras and is compatible with state-of-the-art digital and mixed-signal neuromorphic processing systems. The response of the visual motion pathway of our network resembles the activity of motion-sensitive neurons in the visual system of flies. We ran closed-loop experiments with an autonomous agent in a variety of conditions to assess the collision avoidance and gap finding capabilities of our network. These conditions were chosen from the biological evidence for collision avoidance obtained for flying insects (empty box \cite{Schilstra1999}, corridors \cite{Serres2008, Baird2005, Kern2012blowfly, Linander2015}, gap crossing \cite{Baird2016, Ravi2019,Ravi2020}, and cluttered environments \cite{Mountcastle2016}).
%Gonsek2020
Our agent, utilising its underlying neural network, manages to stay away from walls in a box, centres in corridors, crosses gaps and meanders in cluttered environments. Therefore, it may find applications for autonomous vehicles. Besides, it may serve as a theoretical playground to understand biological systems by using neuromorphic principles replicating an entire action-perception loop.

\section{Results}
\label{sec:results}

\begin{figure}[t]
    \centering
    \includegraphics[width=\textwidth]{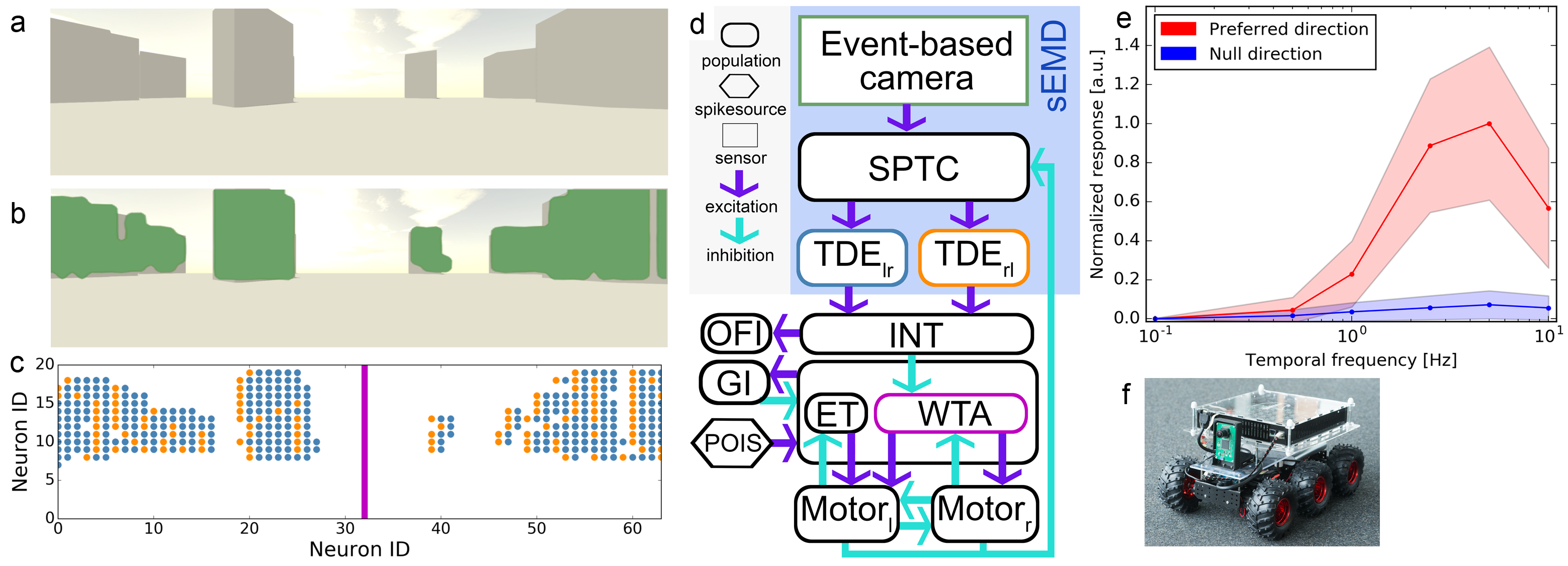}
    \caption{(a-c) Network response in a cluttered environment, (d) collision avoidance network, (e) normalised \ac{sEMD} mean response to a square wave grating and (f) robot used in real-world experiment. a) Cluttered neurorobotics platform environment. The obstacle walls are covered with vertical square-wave-gratings only visible for the event-based camera b) Green areas: simulated event-based camera events directly extracted from the Neurorobotics visual front-end while the agent is slowly moving through the scene in a). c) Bright blue and orange dots: \ac{TDE} left-right and right-left spike response to the scene in a) binned over \texttildelow0.5 seconds of simulation time, pink stripe: inverse \ac{WTA} spike which indicates an obstacle free direction. This spike activates the agent's next turning movement (saccade) by exciting one of the two motor populations in d). The location of the winning neuron defines the direction and duration of the upcoming turn.
    d) Collision avoidance network with \acfp{sEMD} which consist of an event-based camera, a \acf{SPTC} population and two \acf{TDE} populations. Event-based camera (sensory input), \ac{SPTC} population (noise filter and downsampling), \ac{TDE} populations (time-to-travel translation to spike rate and \acf{ISI}), \acf{INT} population (reduces 2D retinotopical map to 1D), inverse \acf{WTA} population (detects minimum of \ac{OF}, hence obstacle free direction), \acf{ET} population (triggers turn when inverse \ac{WTA} can not find direction), \acf{MOT} populations (control turn direction and duration), \acf{OFI} population (modulates robot velocity), \acf{POIS} (drive decision process with Poisson spike process) and \acf{GI} population (suppresses loosing neurons in inverse \ac{WTA} population and \ac{ET} population). e) Normalised preferred direction and null direction mean response of two Nest \ac{sEMD} populations to a square wave grating moving in one cardinal direction with a wavelength of \SI{20}{\degree} and \SI{100}{\percent} relative contrast recorded with an event-based camera at \SI{5000}{\lux} illumination. The standard deviation was calculated on the response of the \ac{sEMD} population. f) Robot used for real-world experiment in Figure \ref{fig:behaviours}d. An embedded event-based camera serves as input to a SpiNN-5 board which drives the motor controllers through an Odroid mini-computer.}
    \label{fig:network_response}
\end{figure}

The \ac{SNN} model proposed in this work consists of two main components, namely a retinotopical map of insect-inspired motion detectors, i.e. \acp{sEMD} \cite{Milde2018sEMD}, and an inverse soft \ac{WTA} network (see Figure \ref{fig:network_response}d and Methods Figure \ref{fig:gap_network}). The former extracts \ac{OF} which, during a translation, is anti-proportionally related to the agent's relative distance to objects in the environment. The latter searches for a region of low apparent motion, hence an obstacle free direction (see Figure \ref{fig:network_response}a-c). After the detection of such a path in the environment the agent executes a turn towards the new movement course. We characterised the network in two steps. First we evaluated the \ac{sEMD}'s response and discussed similarities to its biological counterpart, i.e. T4/T5 neurons, which are thought to be at the core of elementary motion processing in fruit flies \cite{arenz2017temporal, drews2020dynamic}. Second, to further prove the real-world applicability of \ac{sEMD} based gap finding in an \ac{SNN}, we performed closed-loop experiments. We simulated an agent seeing the world through an event-based camera in the Neurorobotics physical simulation platform \cite{falotico2017NRP}. The camera output was processed by the \ac{SNN} resulting in a steering command. We selected a set of parameters that yield the agent to keep at least a mean clearance of \texttildelow6 a.u.\footnote{A.u.: Arbitrary unit, distance divided by robot size, see section \nameref{subsection:environments}} to objects in a box and to enter corridors only with a width greater than 10 a.u. (see Appendix section \nameref{sec:appendix_connectivity}). We tested the performance of this simulated agent with these parameters in all reported experimental conditions hereafter. These experimental conditions were inspired by previous experiments with flying insects.

\subsection{Spiking Elementary Motion Detector}
\label{subsec:results sEMD}

The \ac{sEMD} represents an event-driven adaptation for neuromorphic sensory-processing systems of the well established correlation-based elementary motion detector \cite{hassenstein1956emd}. To evaluate the response of the \ac{sEMD} in the Nest simulator \cite{Diesmann2003NESTAE}, we compared the normalised velocity tuning curves of its ON-Pathway (with recorded event-based camera's input) to the corresponding normalised tuning curve of \textit{Drosophila's} T4 and T5 neurons \cite{Haag2016}. Both velocity tuning curves are determined in response to square-wave gratings with \SI{100}{\%} contrast and a wavelength of \SI{20}{\degree} moving at a range of constant velocities (with temporal frequencies from \SI{0.1}{\Hz} to \SI{10}{\Hz}). The \ac{sEMD} preferred direction exhibits a bell-shaped velocity tuning curve (see Figure \ref{fig:network_response} e), which has the maximum response (mean population activity) at \SI{5}{\Hz}. The null direction response is much lower than the preferred direction. 

The \ac{sEMD} model, which is composed of an event-based camera, a \acf{SPTC} population and the \acf{TDE} (see Figure \ref{fig:gap_network}), exhibits a drop in its output response when the temporal frequency exceeds \SI{5}{\Hz}.
This drop is, however, not anticipated from the \ac{TDE}'s response (see Figure \ref{fig:sEMD}). We would expect the response to saturate at high temporal frequencies since the \ac{TDE} produces interspike intervals and firing rates proportional to the time difference between the two inputs of the \ac{TDE}. The drop in response being rather a consequence of the motion detector model itself, we suggest it to be a consequence of the spatio-temporal band-pass filtering installed by the \ac{SPTC} layer.
While low temporal frequencies lead to unambiguous spatio-temporally correlated and causal \ac{SPTC} spikes from adjacent neurons, high temporal frequencies lead to anti-correlated and non-causal spikes.
Thus, the \ac{TDE} can no longer (spatially) match the spikes unambiguously, which results in a bell-shaped velocity tuning curve of the preferred direction response. 

A similar bell-shaped velocity tuning curve can be observed in \textit{Drosophila}'s T4 cells \cite{Haag2016,arenz2017temporal, Borst2019}. While \textit{Drosophila}'s velocity tuning curves peak at \SI{3}{\Hz} in a drug induced flying state, the \ac{sEMD}'s preferred direction velocity tuning curve peaks at \SI{5}{\Hz}. This suggests that based on the reported parameter set of the \ac{sEMD}, it is tuned to higher relative velocities. The model performs in a robust way for a wide range of illuminations (from \SI{5}{lux} to \SI{5000}{lux}) and relative contrasts (\SI{50}{\percent} response reached at approximately \SI{35}{\percent} relative contrast), as shown in Figure \ref{fig:IlluminationAndContrast}. The \ac{sEMD} approximates the elementary motion processing in the fly brain. This processing is part of the input to the flight control and collision avoidance machinery, hence it can be used as an input for determining a collision-free path.

\subsection{Agent's Behaviour}
\label{subsec: results agent}
The robot's collision avoidance performance was evaluated in an experiment with the agent moving through environments with varying obstacle density. To further understand the mechanisms underlying the robot's movement performance two more experiments were designed. The agent's gap crossing behaviour and tunnel centering behaviour were investigated. 
These behaviour were analysed in insects in a plane, therefore little is known about the effect of flying altitude in most behaviour. We limited our agent to a 2D motion due to this limited understanding. 

\begin{figure}[t]
    \centering
    \includegraphics[width = \textwidth]{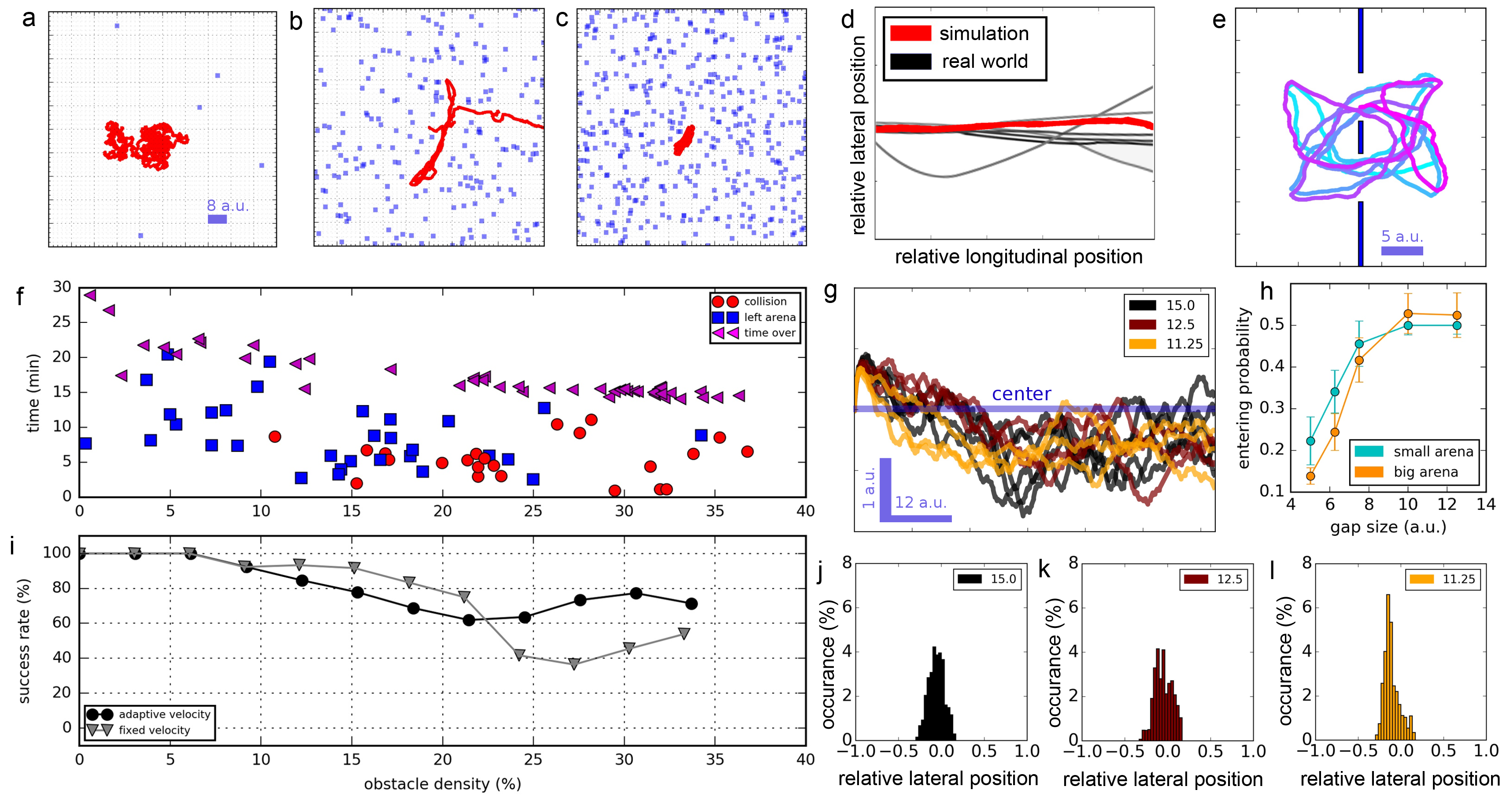}
    \caption{Agent's behaviour in different environments. a-c) Trajectories recorded in arenas with increasing obstacle densities. d) Comparison of real-world centering behaviour (red) to Neurorobotics Platform behaviour (black) in a corridor with normalised corridor width and an absolute corridor length of approximately one meter. e) Simulated robot's trajectory in the gap crossing experiment in a large arena. Colour represents time ($t_0$: light blue, $t_{end}$: magenta). f) Simulated robot's performance in different environments as shown in a-c with modulated velocity. Simulation time at which the simulated robot leaves the arena, collides or the time is over. g) Trajectories in tunnels with a tunnel width of 15, 12.5 and 11.25 a.u.. h) Gap crossing probability in dependency of the gap width for a large and a small arena. i) Simulated robot's performance in cluttered environments as shown in a-c with modulated velocity (black, calculated from data in f) and fixed velocity (grey). Agent's success rate, hence number of runs without collisions.  j-l) Agent's variance from tunnel center for different tunnels.}
    \label{fig:behaviours}
\end{figure}

\subsubsection{Densely Cluttered Environments}
\label{subsubsec: cluttered}

We evaluated the agent's collision avoidance performance in an arena with an obstacle density\footnote{Obstacle density: Percentage of total area covered with objects.} between \SI{0} and \SI{38}{\percent}(0.05 objects per square a.u.). The simulation stops either when the robot collides with an obstacle\footnote{Collision: Simulated robot's outline overlaps with area occupied by object}, when it leaves the arena, or when the simulation real-world-time of six hours is over (see Figure \ref{fig:behaviours}f). At low obstacle densities ($<5\%$) there exist several collision-free paths. The robot exhibits a random walk as the decision making inverse \ac{WTA} neuron population is receiving background spiking activity sampled from a Poisson process (see Figure \ref{fig:behaviours}a,f). In the absence of \ac{OF} input the Poisson distributed background spikes dominates the inverse \ac{WTA} output which results in a probabilistic decision process. The decisions made by the network become less probabilistic with increasing obstacle density since the robot starts to follow the locally low object-density paths forming in the environment (see Figure \ref{fig:behaviours}b,f). At obstacle densities higher than \SI{20}{\percent} most of the gaps in the environment are smaller than the robot's minimum mean obstacle clearance\footnote{Obstacle clearance: Robot's distance to the center of the closest object.} of 7 a.u. (see Figure \ref{fig:clearance} left) so that the agent stays close to its start location (see Figure \ref{fig:clearance} right and Figure \ref{fig:behaviours}c,f). In this range the robot starts to crash into obstacles reaching a minimum success rate of around \SI{60}{\percent} at \SI{22}{\percent} obstacle density. For higher obstacle densities the success rate increases again (see Figure \ref{fig:behaviours}i). A collision of the robot is generally caused by the robot's too long reaction time in an environment with low mean obstacle clearance, hence with high obstacle density (see Figure \ref{fig:clearance}). Since the robot only senses visual stimuli in a 140 degrees horizontal visual field, symmetrically centered around its direction of motion, there is a blind-spot behind the agent. After a strong turn the simulated robot might be confronted with a previously not seen object and directly crash into it. Nevertheless, the agent shows a very robust gap centering behaviour in a large range of different environments with obstacle densities between \SI{0} and \SI{38}{\percent}. The robot's mean success rate amounts to \SI{81}{\percent}.
\\
\indent While local \ac{OF} is instrumental in finding gaps, global \ac{OF} provides information about the clutteredness of the environment. Flies and bees decrease their flying speed when the clutteredness of the environment increases \cite{Kern2012blowfly, Baird2005}. Our agent regulates its speed based on the global \ac{OF} and, consequently, moves slower in denser regions of the environment (see Figure \ref{fig:velocity}). To examine the effect of the velocity dependency, we ran a second experiment with the robot moving with constant velocity (see Figure \ref{fig:behaviours}i and Figure \ref{fig:cluttered}). With velocity control collisions were encountered only in few runs, however, for obstacle densities higher than 24 percent the number of collisions significantly increased when the velocity was kept constant.

\subsubsection{Gaps}
\label{subsubsec: results gaps}

When presented with a choice between two gaps of different size bees prefer to pass the larger gap \cite{Baird2016, Ong2017}. This behaviour decreases the insect's collision probability significantly. While bees might choose the gap in a complex decision process \cite{Ravi2020} our agent's preference underlies a simple probabilistic integration mechanism. The simulated robot's upcoming movement direction is determined by an inverse \ac{WTA} spike occurring in an obstacle-free direction as shown in Figure \ref{fig:network_response}a-c. When confronted with a small and a large gap the probability of an inverse \ac{WTA} spike appearing in the greater gap is higher. Hence, we assume that the robot automatically follows pathways with a larger gap size. To evaluate this assumption we observed the robot's gap crossing in an arena with two alternative gaps (see Figure \ref{fig:behaviours}e). 
The robot can decide to cross any of the two gaps or stay in one half of the arena. There is a competition between staying in the open-space and crossing a gap. The larger the gap size is, the more likely the robot will cross a gap. We investigated the probability to cross gaps by having two gaps, one with a fixed gap size (10 times the agent width), the other with a gap size between 5 a.u and 13 a.u.
We calculated the gap entering probability by comparing the number of passes through both gaps. As expected the entering probability increases with gap size until a width of 10 a.u. (see Figure \ref{fig:behaviours}h). For a larger gap width the entering probability does not change significantly. However, for smaller gap sizes the probability of a spike pointing towards open space in the inverse \ac{WTA} becomes significantly higher. Therefore, the robot prefers to pass through gaps of larger size. Besides the gap width the arena size changes the passing probability. In a smaller arena the simulated robot stays closer to the gap entry which increases the relative gap size sensed by the agent. Therefore, a larger part of the vehicle's visual field is occupied by the gap entry which increases the probability of a spike occurring in the gap area. In a smaller arena we observed that the robot's gap entering probability is higher for gaps smaller then 10 a.u. than in a big arena (see Figure \ref{fig:behaviours}h). A decrease in arena size can be compared to an increase in obstacle density since both parameters reduce the robot's obstacle mean clearance (see Figure \ref{fig:clearance}, left). Therefore, the agent tends to enter gaps of smaller size in densely cluttered environments. This automatic scaling mechanism keeps the agent's collision probability very low in sparsely cluttered environments by staying away from small gaps. In environments with high obstacle density the robot still keeps its mobility by passing through smaller gaps. Finally, when the obstacle density exceeds \SI{20}{\percent}, most gaps fall below the gap entering threshold so that the robot can not leave the arena anymore (see Figure \ref{fig:clearance}, right and Figure \ref{fig:behaviours}c,f).

\subsubsection{Corridors}
\label{subsubsection:resultscorridors}
One common experiment to characterise an agent's motion vision response is to observe its centering behaviour in a tunnel equipped with vertical stripes on the walls. The simple geometry of the environment enables the observer to directly relate the received visual input with the agent's actions. In bees and flies an increase in flight velocity proportionally to the tunnel width has been observed \cite{Kern2012blowfly, Lecoeur2019TheRO, Baird2005}. In very narrow tunnels insects show a pronounced centering behaviour which declines with increasing tunnel width. We evaluated the robot's performance in three tunnels with different tunnel widths. Similar to the biological role model the robot's velocity stands in a positive linear relationship with the tunnel width. The measured velocity in a.u. per second is \texttildelow0.79, \texttildelow0.75 and \texttildelow0.72 for a tunnel width of 15, 12.5 and 11.25 a.u. respectively. Furthermore, the robot always stays in the center of the tunnel, especially in very narrow tunnels (see Figure \ref{fig:behaviours}g). The deviation from the tunnel center is proportional to the tunnel width (for the simulated robot, see Figure \ref{fig:behaviours}j--l, physical robot see \ref{fig:behaviours}d). Therefore, similar to observations in blowflies, the robot's lateral position in the tunnel changes linearly with the tunnel width \cite{Kern2012blowfly}.

\section{Discussion}
\label{sec:discussion}
Autonomous agents need to successfully avoid obstacles in a variety of different environments, be they human made or of natural origin. Our investigations present a closed-loop proof of concept of how obstacle avoidance could be performed in a parsimonious, asynchronous and fully distributed fashion. While most results reported here are based on computer simulations, the implementation on digital or mixed-signal neuromorphic hardware of each building block of the simulated \ac{SNN} have been demonstrated for event-based cameras \cite{Lichtsteiner2008}, the \ac{sEMD} \cite{Milde2018sEMD, Schoepe2019sensoryintegration} (see Figure \ref{fig:IlluminationAndContrast}), artificial neurons \cite{Indiveri_etal06} and synapses \cite{Bartolozzi_Indiveri07}, as well as the inverse \ac{WTA} \cite{horiuchi2009batnavigation}. We demonstrated for the first time a simulation of a neuromorphic system that takes informed decisions while moving in its environment by closing the action-perception loop. We emulated this system on neuromorphic sensory-processing hardware carried by a physical robot (see Figure~\ref{fig:network_response}f, ~\ref{fig:behaviours}d,~\ref{fig:robot} and~\ref{fig:real_worl_results}), tested it in a corridor centering experiment, and obtained similar results to the simulation. These real-world experiments suggest that the underlying computational primitives lead to robust decision making in operational real-time. 
Due to the physical simulation with the engine Gazebo that capture the physics of the movements and our real-world proof of implementation, our simulations are likely to translate to real-world situations. 
While producing relatively simple, yet crucial decisions, the proposed model represents a critical milestone towards enabling parallel, asynchronous and purely event-driven neuromorphic systems. 
  
Our proposed \ac{SNN} architecture comprises of \texttildelow 300k synapses and \texttildelow 4k neurons which yields a low-power, lightweight and robust neural algorithm. When implemented on mixed-signal neuromorphic processing hardware, e.g. \cite{Moradi_etal17, painkras2013spinnaker, Wang_vanSchaik18}, the payload required to perform on-board processing will be drastically reduced. This reduction stems from the low volume and lower power requirements of neuromorphic hardware. In addition such hardware implementation would ensure operational real-time decision making capabilities.
The features outlined above are quite desirable in the context of highly constrained autonomous systems such as drones or other unmanned vehicles.
\\
\indent We investigated the performance of the \acp{sEMD}, the apparent motion encoders in our \ac{SNN}, in detail. The \acp{sEMD} show a similar velocity response curve to motion-sensitive neurons (e.g. T4 and T5 neurons in the fruitfly's brain \cite{Borst2019, Mauss2020optic_flow}) when presented with a grating of \SI{20}{\degree} spatial frequency and temporal frequencies between \SI{0.1} and \SI{10}{\Hz}. Due to the logarithmic illumination sensitivity of the event-based cameras the motion vision apparatus is very robust against absolute brightness levels in the range of \SI{5} up to \SI{5000}{lux}. The \ac{sEMD} model shows a much higher sensitivity regarding contrast changes than its biological role model. Current research suggest that \textit{Drosophila}'s optical lobe performs contrast normalisation through inhibitory recurrent feedback to evoke a contrast independent response \cite{drews2020dynamic}. In a next step we will implement contrast normalisation in our motion vision network to improve its performance in natural environments. 
\\
\indent
Besides the similarities in neural response, the agent showed many similarities to flying insects in its behaviour in spatially constrained environments. 
It meandered in cluttered terrain (Section \nameref{subsubsec: cluttered}), modulated its speed as a function of object proximity (Section \nameref{subsubsection:resultscorridors}), selected wider gaps (Section \nameref{subsubsec: results gaps}), centered in tunnels (Section \nameref{subsubsection:resultscorridors}), while using an active gaze strategy known as saccadic flight control (Section \nameref{subsection:GFnetwork})\cite{Kern2012blowfly, Lecoeur2019TheRO, Baird2005, Ong2017, Schnell2017ADN}. 
The agent moved collision-free through cluttered environments with an obstacle density between \SI{0} and \SI{38}{\percent} with a mean success rate of \SI{81}{\percent}\footnote{Several closed-loop, insect-inspired approaches have been demonstrated \cite{Fu_etal19, Serres_Ruffier17}, however, due to a missing unifying benchmark and evaluation metric, to compare insect-inspired collision avoidance algorithms, we cannot provide a quantitative comparison}. 
We further examined the simulated robot's performance to understand the essential behavioural components which led to a low collision rate. 
The most significant ingredient in that regard was the implementation of an \ac{OF} strength dependent locomotion velocity.
This insect inspired control mechanism improved the collision avoidance performance of the agent from a mean success rate of \SI{76}{\percent} to \SI{81}{\percent} (Compare Figure \ref{fig:behaviours}i and Figure \ref{fig:cluttered}). We propose that this velocity adaptation mechanism could be regulated in insects by a simple feedback control loop. This loop changes the agent's velocity anti-proportionally to the global \ac{OF} integrated by a subset of neurons (For further explanations see Section \nameref{subsection:GFnetwork}).
\\
\indent An \ac{OF}-dependent control of locomotion velocity is only one of at least three mechanisms which decreased the agent's rate of collision. When moving in environments of high obstacle density the simulated robot follows locally low obstacle density paths. We suggest that a probabilistic decision process in the network model automatically keeps the agent's collision probability low by following these pathways. We further investigated this path choice mechanism in a second experiment. Here, the agent had to cross two gaps of different size. The dependence of the agent' probability to cross the gap resembled that of bees \cite{Ong2017}. Similar to insects the agent preferred gaps of larger size. 
Bees cross gaps with a gap-size as small as 1.5 times their wingspan \cite{Ravi2020}. In contrast our agent crossed gaps of 5 times its body width. This discrepancy in performance may be due to the absence of a goal. A goal can be understood as providing an incentive to cross a gap despite a risk of collision. Indeed in behavioural experiments, bees had to cross the gap to return to their home. Combining different directions, such as a collision-free path and a goal, require an integration of the two signal representations. 
Such networks have been proposed for navigating insects \cite{Sun_etal20}. Integration of similar streams of information have been demonstrated to work in neuromorphic systems \cite{Kreiser_etal18b,Schoepe2019sensoryintegration, Blum_etal17}, however, we envision that a dynamic competition between collision avoidance and goal reaching neural representations could allow our robot to cross gaps 1.5 times its width.

The findings reported here indicate an alternative point of view how flies and bees could use motion-vision input to move through the environment, not by collision avoidance but by gap finding. As also stated by Baird and Dacke \cite{Baird2016}, flies and bees might not actively avoid obstacles but fly towards open space, i.e. gaps. Looking at our network, we suggest that \ac{WTA} alike structures in flying insect brains might integrate different sensory inhibitory and excitatory inputs with previously acquired knowledge to take navigational decisions. One could think of the central complex as such a structure which has been described recently in several insect species \cite{Honkanen2019CXreview}. 
\\
\indent The third mechanism is the agent's centering behaviour. By staying in the middle of a tunnel with similar patterns on both walls the simulated robot minimises its risk of colliding with a wall.
The agent's deviation from the tunnel center changes approximately linearly with the tunnel width. These results show a very strong resemblance with experimental data from blowflies (see Figure \ref{fig:behaviours}j--l) \cite{Kern2012blowfly}. So far centering behaviour was suggested to result from balancing the \ac{OF} on both eyes. Centering in a tunnel can be seen as crossing elongated gaps.
Our agent is also able to cross gaps. Two hypothesis have been suggested to cross gaps in flying insects, using the \ac{OF} contrast \cite{Ravi2019} and the brightness \cite{Baird2016}. Our results suggest that collision avoidance could be mediated by identifying minimum optic flow to center in tunnel, cross gaps, or meander in cluttered environment. This strategy has so far not been investigated in flying insects. The main hypothesis to control flight in clutter is to balance either an average or the maximum \ac{OF} on both eyes \cite{Lecoeur2019TheRO}. Further behavioural experiments are required to disentangle between the different strategies and their potential interaction. Building on the work of \cite{Baird2016}, the different hypothesis could be placed into conflict by creating a point-symmetric \ac{OF} around the gap center (leading to centering), a brightest point away from the gap center, and a minimum \ac{OF} away from the center (e.g. by using an \ac{OF} amplitude following a Mexican hat function of the radius from the geometric center). 
\\
\indent Our model shares several similarities with the neural correlate of visually-guided behaviour in insects, including motion-sensitive neurons \cite{Mauss2020optic_flow}, an integration of direction \cite{Sun2020navigation}, efference copy to motion-sensitive neurons \cite{Kim2015CellularEF}, and neurons controlling the saccade amplitude \cite{Schnell2017ADN}. Our agent was able to adopt an active gaze strategy thanks to a saccadic suppression mechanism (due to an inhibitory efference copy from the motor neurons to the inverse \ac{WTA} and motion-sensitive neurons). 
When the inverse \ac{WTA} layer did not "find" a collision-free path (i.e. a solution to the gap finding task), an alternative response (here a U-turn) was triggered thanks to global inhibitory neurons and excitatory-inhibitory networks (GI-WTA-ET, for more details see Section~\nameref{subsection:GFnetwork}). 
The neuronal correlate of such a switch, to our knowledge, has not been described in flying insects. Our model, thus, serves as a working hypothesis for such a neuronal correlate. Furthermore, by varying the connection between \ac{sEMD}-inverse \ac{WTA}, we could allow the agent to cross smaller gaps. We hypothesise that differences in clearance or centering behaviour observed between insect species \cite{Baird2016, Ravi2019} could be due to different wiring or modulation between motion-sensitivity neurons and direction selection layer, likely located in the central complex.
\\
\indent In this study we demonstrated a system-level analysis of a distributed, parallel and asynchronous neural algorithm to enable neuromorphic hardware to perform insect-inspired collision avoidance. To perform a wide variety of biological-relevant behaviour the network comprised approximately 4k neurons and 300k synapses. The agent guided by the algorithm robustly avoided collision in a variety of situations and environments, from centering in a tunnel to crossing densely cluttered terrain and even gap finding, solved by flying insects. These behaviour were accomplished with a single set of parameters, which have not been optimised for any of those. From the investigation of the agent and its underlying behaviour, we hypothesise that insects control their flight by identifying regions of low apparent motion, and that excitatory-inhibitory neural structures drive switches between different behaviours. With these investigations we hope to advance our understanding of closed-loop artificial neural computation and start to bridge the gap between biological intelligence and its neuromorphic aspiration. 

\newpage

\section{Methods}
\label{sec:methods}

Most experiments in this article were conducted in simulation using either the Nest \acf{SNN} simulator \cite{Diesmann2003NESTAE} or the Neurorobotics Platform environment \cite{falotico2017NRP}. A corridor centering experiment was conducted in a real-world corridor centering experiment using a robotic platform equipped with the embedded Dynamic Vision Sensor\footnote{The embedded Dynamic Vision Sensor follows the same operational principles of event-based cameras as described in Section~\nameref{sec:vision} but features a much more compact design} as visual input and a SpiNN-5 \cite{painkras2013spinnaker} board for \ac{SNN} simulation in computational real-time. Sensory data for the \ac{sEMD} characterisation were recorded with an event-based camera in a real world environment. The hardware, software, \ac{SNN} models and methodologies used in this article are explained in the following. 

\subsection{Spiking Neural Networks}
\label{subsec:meth SNNs}
In contrast to conventional processing as postulated by von Neumann \cite{Laird2009} which is characterised by synchronous and inherently sequential processing, neural networks, whether rate-based or spike-based, feature parallel and distributed processing.
Artificial neural networks, the rate-based counterpart of \acp{SNN}, perform synchronous and clock-driven processing, \acp{SNN}, additionally, feature an asynchronous and event-driven processing style.
\acp{SNN} represent a promising alternative to conventional von Neumann processing and hence computing which potentially feature low-latency, low-power, distributed and parallel computation.
Neuromorphic hardware present a solution to the aforementioned limitations of conventional von Neumann architectures including parallel, distributed processing \cite{thakur2018neuromorphic} in the absence of a central clock \cite{Mead_1989, Liu_2015}, as well as co-localisation of memory and computation \cite{Payvand_etal19, Serb_etal20}.
Moreover, neuromorphic processors benefit from the underlying algorithm to be implemented in a \ac{SNN}.
Emulating a \ac{SNN} on a neuromorphic processor (especially a mixed-signal one) enables the network to operate in continuous time\footnote{A time-continuous mode of operation, in contrast to a time-varying one, is characterised by the absence of a fixed sampling frequency} as time represents itself \cite{Mead_1989}.
\acp{SNN} consist of massively parallel connected networks of artificial synapses and spiking neurons \cite{gerstner_kistler_2002}. 
\acp{SNN}, as any processing algorithm, aim to structure and represent incoming information (e.g. measurements) in a stable, robust and compressed manner (e.g. memory).
Measurements sampled at fixed time intervals have the disadvantage that collected data is highly redundant and prone to aliasing if the signal of interest varies faster than half the sampling frequency.
Event-driven approaches to sampling alleviate these limitations.
As incoming measurements shouldn't be sampled at fixed temporal intervals, they need to be taken based on fixed or relative amplitude changes to take full advantage of the time-continuous nature of \acp{SNN} and neuromorphic hardware.
Such measurements can be obtained from different sensory domains (e.g. touch \cite{Khalil2017CMOS}, smell \cite{drix2020resolving}, auditory \cite{Liu2014AsynchronousBS, vanSchaik2007AEREAR} and vision \cite{Mahowald1992VLSI, Lichtsteiner2006}), with vision being the most studied and well understood sensory pathway (but see \cite{Olshausen2005How} for a critical review) both in the  brain and its artificial aspiration.
While images taken with conventional cameras can be converted to spike trains which are proportional to the pixel intensity\footnote{To perform this conversion one can use a different encoding schemes including rank-order code \cite{Thorpe1998Rank}, timing-code \cite{Thorpe1990SpikeAT}, \cite{Masquelier12} or Poisson rate-code.}, event-based cameras directly sample only relative changes of log intensity and transmit events.
A variety of event-based cameras have been proposed in the last two decades \cite{Lichtsteiner2006, Lichtsteiner2008, Posch2010} that all feature an asynchronous, parallel sampling scheme\footnote{Level sampling means that a given time-continuous signal is sampled when the level changes by fixed (relative) amount $\epsilon$, whereas time sampling, i.e. Nyquist-Shannon sampling, means that the signal is sampled when the time has changed by fixed amount $\epsilon$} in which changes are reported at the time of occurrence in complete time-continuous manner. 
The output of event-based cameras is hence ideally suited to be processed by an \ac{SNN} implemented on a neuromorphic processor.
We collected real-world data using the DVS128 event-based camera \cite{Lichtsteiner2008} to characterise the \ac{sEMD} response (see Figure \ref{fig:network_response}e). The event-based camera comprises 128$\times$128 independently operating pixels which respond to relative changes in log-intensity, i.e. in temporal contrast. When the change in light intensity exceeds an adaptive threshold the corresponding pixel produces an event. The address and polarity of the pixel are communicated through an Asynchronous Event Representation bus \cite{Mahowald1992VLSI}. Light increments lead to ON-events, whereas light decrements lead to OFF-events. The sensor reaches a dynamic range of more than 120 dB and is highly invariant to the absolute level of illumination due to the logarithmic nature of the switched-capacitor differencing circuit \cite{Lichtsteiner2006, Lichtsteiner2008}. \\

\subsection{Spiking Elementary Motion Detector}
\label{subsec: meth sEMD}

\begin{figure}[th!]
\centering
\includegraphics[width = 0.5\textwidth]{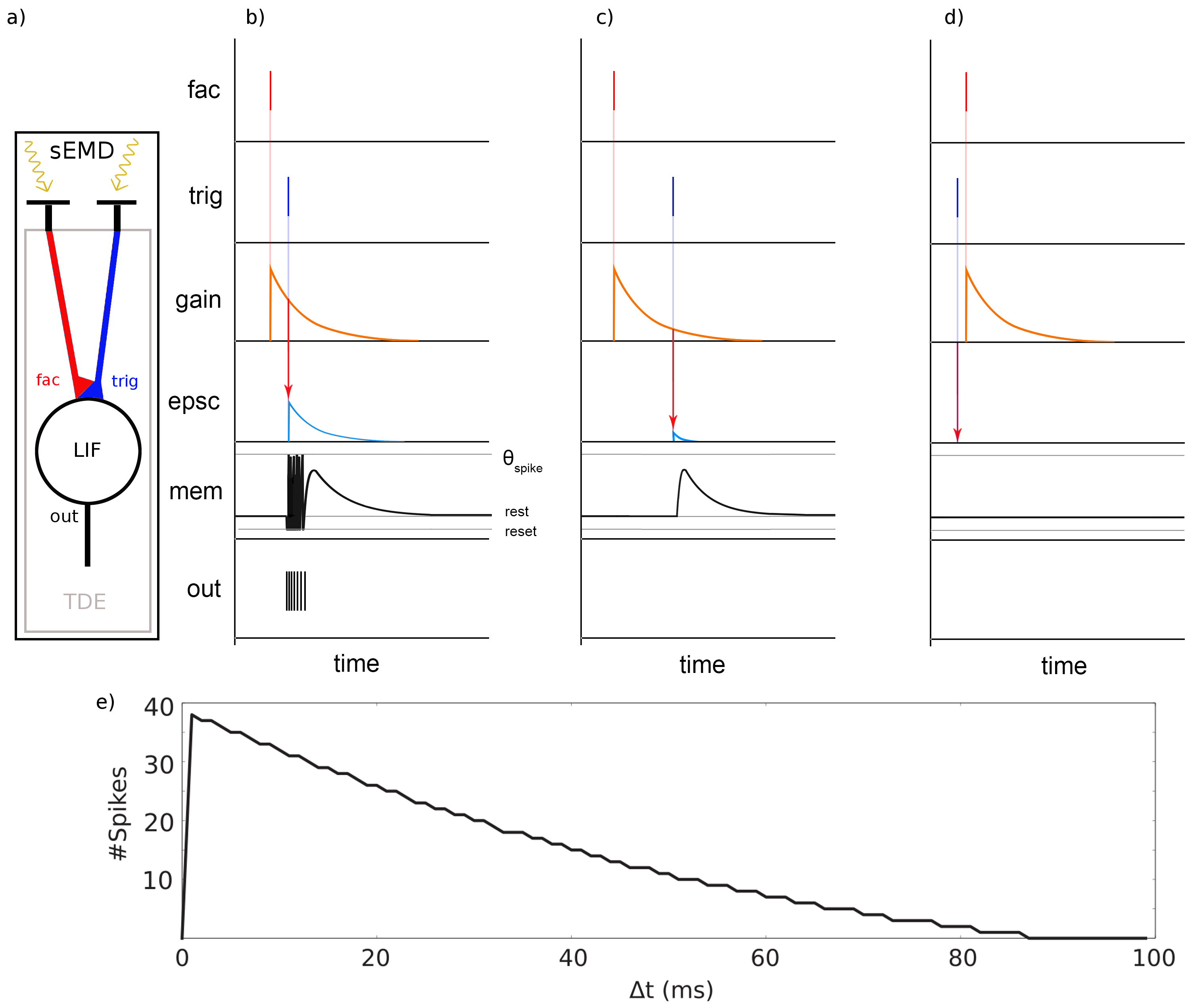}
\caption{Spiking Elementary Motion Detector model adapted from \cite{DAngelo2020EventBasedEM}. a) \ac{sEMD} model consisting of visual input and \ac{TDE} unit. Two adjacent retina inputs are connected to the facilitatory synapse (fac) and the trigger synapse (trig). The fac synapse controls the gain of the trig synapse postsynaptic current (epsc) which integrates onto the \ac{LIF} neuron's membrane potential which produces output spikes (out). b) model behaviour for small positive $\Delta t$. c) behaviour for large positive $\Delta t$. d) behaviour for negative $\Delta t$. e) number of output spikes over $\Delta t$.}
\label{fig:sEMD}
\end{figure}

In 2018 we proposed a new insect-inspired building block for motion vision in the framework of \acp{SNN} designed to operate on the out event-stream of event-based cameras, the \ac{sEMD}\cite{Milde2018sEMD}. The \ac{sEMD} is inspired by the computation of apparent motion, i.e. \acf{OF}, in flying insects. In contrast to its correlation-based role model the \ac{sEMD} is spike-based. It translates the time-to-travel of a spatio-temporally correlated pair of events into direction dependent, output burst of spikes. 
While the \ac{sEMD} provides \ac{OF} estimates with higher precision when the entire burst is considered (rate-code), the interspike interval distribution (temporal-code) within the burst provides low-latency estimates.
The \ac{sEMD} consists of two building blocks, a retina to extract visual information from the environment, and the \ac{TDE} which translates the temporal difference into output spikes (see Figure \ref{fig:sEMD}a). When the \ac{sEMD} receives an input spike at its facilitatory pathway an exponentially decreasing gain variable is generated. The magnitude of the synaptic gain variable during the arrival of a spike at the trigger synapse defines the amplitude of the excitatory post-synaptic current generated. This current integrates onto the \ac{sEMD}'s membrane potential which generates a short burst of output spikes. Therefore, the number of output spikes encodes direction sensitive and anti-proportionally the stimulus' time-to-travel (see Figure \ref{fig:sEMD}e) between two adjacent input pixels. We implemented and evaluated the motion detector model in various software applications (Brian2, Nengo, Nest), in neuromorphic digital hardware (SpiNNaker, Loihi) and also as analog CMOS circuit \cite{Milde2018sEMD, Schoepe2019sensoryintegration}.
%% FPGA paper not accepted so far, do not put citation here so far

\subsection{Collision Avoidance Network}
\label{subsection:GFnetwork}

\begin{wrapfigure}{r}{0.52\textwidth}
    \includegraphics[width=0.5\textwidth]{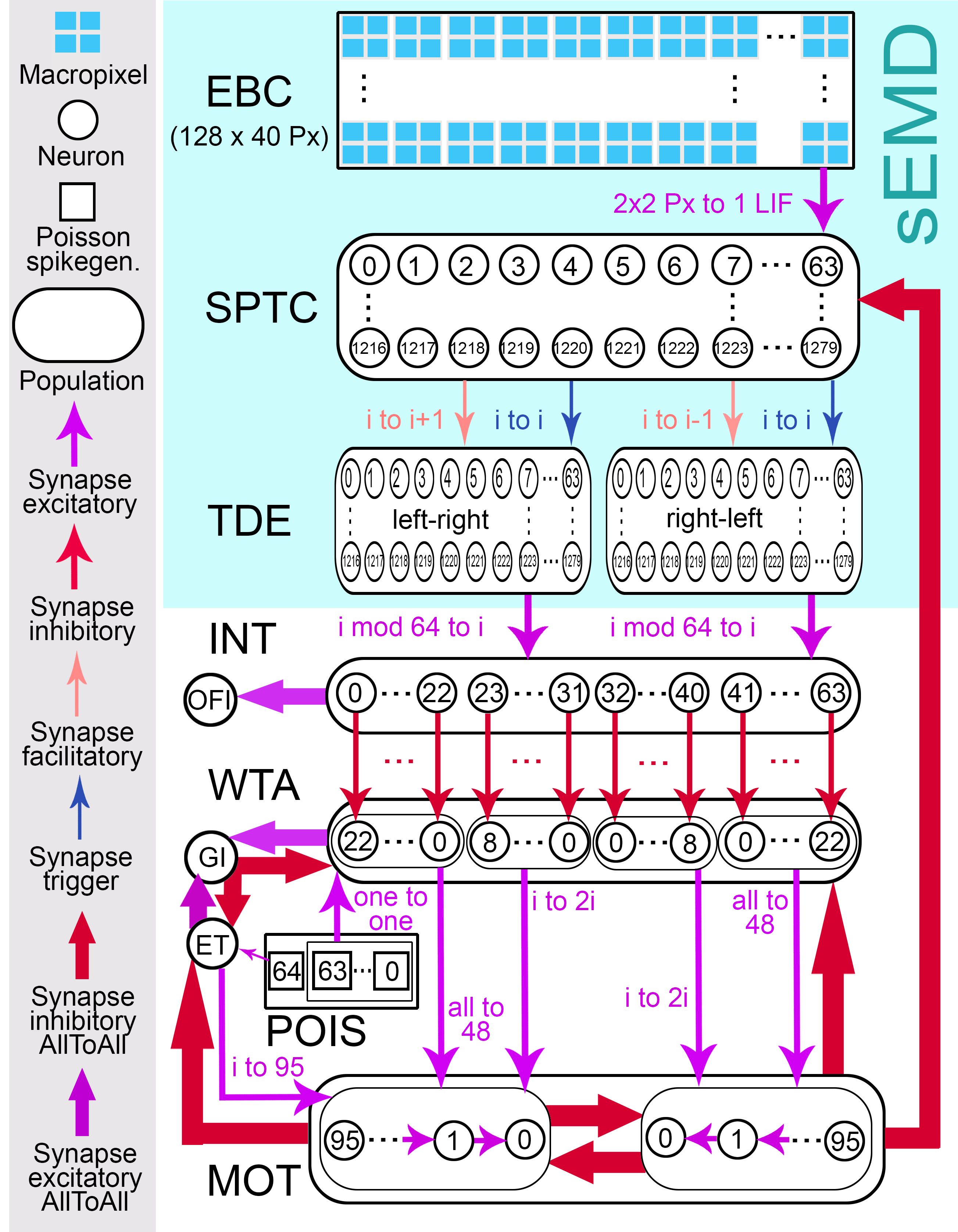}
    \caption{Collision avoidance network. The macropixels (2x2 pixels) of the \acf{EBC} project onto single neurons of the \acf{SPTC} population removing uncorrelated noise. Two adjacent \ac{SPTC} neurons are connected to one \acf{TDE} in the left-right sub-population and the right-left sub-population respectively. Trigger and facilitator connection are opposite in the two populations. The \acf{INT} population reduces the two dimensional retinotopical map to a one-dimensional map by integrating the spikes of each \ac{TDE} column onto a single \ac{LIF} neuron. The inverse \acf{WTA} population and \acf{ET} population become excited by Poisson spike sources. The winner-take-all mechanism is driven by recurrent suppression through the \acf{GI} neuron. The two \acf{MOT} populations are activated by a spike in the inverse \ac{WTA} population. The id of the spiking inverse \ac{WTA} neuron defines which \ac{MOT} becomes activated and for how long. When the \ac{ET} neuron spikes the left \ac{MOT} population becomes activated for the maximal time duration. When the \ac{MOT} population is inactive the robot moves straight foward collecting apparent motion information. When one \ac{MOT} population is active the robot turns. All-to-all inhibition between the \ac{MOT} sub-populations guarantees the dis-ambiguity of the steering commands. Inhibition from the \ac{MOT} to the \ac{SPTC} population suppresses rotational \ac{OF} input which contains no relative depth information. Inhibition from \ac{MOT} to inverse \ac{WTA} hinders the network from taking any new decision during a turn.}
     \label{fig:gap_network}
\end{wrapfigure}

The collision avoidance network (see Figure \ref{fig:gap_network}) extracts a collision-free direction from its \ac{sEMD} outputs and translates this spatial information into a steering command towards open space. The first layer, the event-based camera, generates an event when a relative change in log-illumination, i.e. temporal contrast, is perceived by a pixel. A macropixel consists of 2 x 2 event-based camera pixels. Each macropixel projects onto a single current-based exponential \ac{LIF} neuron (hereafter referred to as \ac{LIF} for sake of clarity) in the \acf{SPTC} layer (in Nest the neuron model used throughout this study is called iaf\textunderscore psc\textunderscore exp). Each single \ac{SPTC} neuron emits a spike only when more than 50\% of the pixels within a macropixel elicit an event within a rolling window of \SI{20}{ms}. Therefore, the \ac{SPTC} population removes uncorrelated events, which can be interpreted as noise. 
Additionally, it decreases the network resolution from 128 times 40 pixels to 64 times 20 neurons. The next layer extracts \ac{OF} information from the filtered visual stimulus. It consists of two \ac{TDE} populations sensitive to the two horizontal cardinal directions respectively.
Each \ac{TDE} receives facilitatory input from its adjacent \ac{SPTC} neuron and trigger input from its corresponding \ac{SPTC} neuron. The facilitatory input might arise either from the left (left-right population) or from the right (right-left population). The \ac{TDE} output encodes the \ac{OF} as number of spikes in a two-dimensional retinotopical map. Since the agent moves on the ground it only estimates the amount of horizontal \ac{OF}. Hence, the subsequent \ac{INT} population integrates the spikes of each \ac{TDE} column in a single \ac{LIF} neuron. This layer encodes the \ac{OF} in a one-dimensional retinotopical map. The subsequent population, an inverse soft \acf{WTA} determines the agent's movement direction, a minimum of \ac{OF} in the one-dimensional retinotopical map. Since \ac{OF} encodes the relative distance to objects during a translational movement this direction represents an object-free pathway, hence the inverse \acf{WTA} is inverted by sending feed-forward inhibition into the neural population. A population of \ac{POIS}  injects Poisson distributed background spikes which ensures a neuron within the inverse \ac{WTA} to win at any moment in time even in the absence of \ac{OF}. In the absence of \ac{INT} input the inverse \ac{WTA} neuron with the strongest \ac{POIS} input wins and suppresses through the \ac{GI} neuron the activity of all others. Local lateral connections in the inverse \ac{WTA} population strengthen the winning neuron due to excitatory feedback (For the sake of clarity recurrent excitation is not shown in Figure \ref{fig:gap_network}). 

Because of the consistently changing nature of the \ac{POIS} spike trains the winner changes frequently and the agent executes a random walk (see Figure~\ref{fig:behaviours}a). 
When the agent approaches an object the position of the obstacle is indicated by a number of spikes in the \ac{INT} population. These spikes strongly inhibit the inverse \ac{WTA} at the corresponding position and its closest neighbours so that this inverse \ac{WTA} direction cannot win. Therefore, the active neurons in the inverse \ac{WTA} always represent an obstacle-free direction. In case no object-free direction has been found for \texttildelow700 milliseconds since the start of an intersaccade the \ac{ET} neuron emits a spike. This neuron is only weakly excited by the \ac{POIS} population and connected to the \ac{GI} neuron similarly to the inverse \ac{WTA} population. Only when the \ac{ET} has not been inhibited for a long time, hence the inverse \ac{WTA} was not able to generate a spike due to strong over all inhibition, the \ac{ET} neuron wins. 
The final layer called \ac{MOT} population translates the inverse \ac{WTA} population and \ac{ET} neuron activity into a turn direction and duration using pulse-width modulation to control the motors. The left turn \ac{MOT} population becomes activated by inverse \ac{WTA} neurons on the left side and the right turn population by inverse \ac{WTA} neurons on the right side. Since the turning velocity is always constant the angle of rotation is defined by the duration of the turn. This duration of the excitation wave in the \ac{MOT} population relates proportionally to the distance of the inverse \ac{WTA} neuron from the center of the horizontal visual field. The duration saturates for neuron distances higher than nine. Since a left turn and a right turn are exclusive events, strong inhibition between the two \ac{MOT} populations assures to disambiguate the \ac{MOT} layer outputs. In case the \ac{ET} neuron emits a spike the excitation wave passes through most neurons of the left \ac{MOT} population. Hence, the turning duration is slightly higher than for any turn induced by the inverse \ac{WTA} population. The agent turns completely away from the faced scene since no collision free path was found in that direction. During the execution of a turn the gap finding network receives mainly rotational \ac{OF}. This type of apparent motion does not contain any depth information and therefore no new movement direction should be chosen during or shortly after a turn. Because of that the \ac{MOT} layer strongly inhibits the inverse \ac{WTA} and \ac{SPTC} populations as well as the \ac{ET} neuron. After a turn has finished and none of the \ac{MOT} populations is spiking anymore the agent moves purely translatory. The movement speed during this phase $v_{ints}$ is defined in equation \ref{eq:velocity agent} where $\bar{f}_{OFI}$ is the mean firing rate of the \ac{OFI} population. During this movement phase, called intersaccade, the agent integrates translational \ac{OF} information in its \ac{INT} population. The inverse \ac{WTA} population slowly depolarizes from its strongly inhibited state and releases a spike indicating the new movement direction. This spike triggers the next saccadic turn of the robot while the id of the winning neuron defines the direction and duration of the movement. 

\begin{equation}
\label{eq:velocity agent}
    v_{ints} (\frac{m}{s}) = 1 - \bar{f}_{OFI} \times 0.001
\end{equation}

\subsection{Neurorobotics Platform}
\label{subsection:NRP}

To perform our behavioural experiments we decided to simulate the entire system, from visual input to actions, using the Neurorobotics Platform.
This platform combines simulated \acp{SNN} with physical realistic robot models in a simulated 3D environment \cite{falotico2017NRP}. 
The platform consists of three main parts, the world simulator Gazebo, the \ac{SNN} simulator Nest and the Transfer Function Manager \ac{BIBI}. The \ac{BIBI} middleware consists of a set of transfer functions which enables the communication between Gazebo and NEST via \ac{ROS} \cite{Quigley2009ROS} and PyNN adapters. The \ac{CLE} synchronizes the two simulators Gazebo and Nest and controls the data exchange through transfer functions. 
The simulation front-end virtual coach is useful to control the whole simulation procedure through a single python script. Furthermore, the State Machines Manager of the SMACH framework can be used to write State Machines which manipulate the robot or world environment during the experiment.

\subsection{Real World Robot}
\label{subsec: real robot}
The robot receives visual input from the embedded Dynamic Vision Sensor with a 60 degrees lens. The event-based camera sends its events to a SpiNN-5 board which simulates a simplified version of the collision avoidance network decribed in the section \nameref{subsection:GFnetwork}. The robot's visual field consists of 128x128 pixels which project onto 32x32 \acp{SPTC}. The robot computes ON-event and OFF-events in two separate pathways from the retina until the \acp{sEMD}. The \ac{INT} layer integrates the spikes of the ON- and OFF-pathway in a single population. The network does not contain any \ac{OFI} neuron and the agent moves with a constant velocity of around 0.5 m/s. There are also no \ac{MOT} populations and no \ac{ET} population. The inhibition from the \ac{MOT} population to the \ac{SPTC} population is replaced by inhibition from inverse \ac{WTA} to \ac{SPTC}. The motor control is regulated on an Odroid mini-computer. The computer receives inverse \ac{WTA} spikes from the SpiNN-3 board via Ethernet and translates these spikes into a motor command which is then send via USB to the motor controller. This causes a long delay between perception and action. The motor controller drives the six-wheeled robot in a differential manner.

\subsection{Event-Based Cameras in Gazebo}
\label{sec:vision}
\indent Kaiser et al. 2016 \cite{kaiser2016DVSNRP} developed a Neurorobotics Platform implementation of an event-based camera based on the world simulator Gazebo. This model samples the environment with a fixed update rate and produces an event when the brightness change between old and new frame exceeds a threshold. We used this camera model in our closed-loop simulations as visual input to the collision avoidance network.
Even though Gazebo produces an event-stream from regularly sampled synchronous frame-difference, our \ac{sEMD} characterisation and open-loop experiments (see Section \nameref{subsection:characterisation} and \cite{Milde2018sEMD}) confirmed the working principle of the motion detector model with real-world event-based camera data.
We could further demonstrate the real-world fully-neuromorphic applicability in closed-loop of most parts of the simulated agent including the apparent motion computation by the \acp{sEMD} and the saccadic suppression \cite{Schoepe2020integration}.
We set the resolution of the Gazebo event-based camera model to 128 times 40 pixels. The reduction of the vertical resolution from 128 to 40 pixels was done to speed up the simulation time and to make the model fit onto a SpiNN-3 board \cite{painkras2013spinnaker}. 
To further accelerate the simulation we limited the number of events per update-cycle to 1000 and set the refresh rate to 200 Hz. Therefore, the \ac{sEMD} can only detect time differences with a resolution of 5 ms. We decided for a large horizontal visual angle of 140 degrees so that the robot does not crash into unforeseen objects after a strong turn. At the same time the uniform distribution of 128 pixels over a 140 degrees horizontal visual field leads to an inter-pixel angle of approximately 1.1 degrees. This visual acuity lies in a biologically plausible range of inter-ommatidial angles measured in Diptera and Hymneoptera which varies between 0.4 and 5.8 degrees \cite{Land97ommatidiae}.

\subsection{Driving Agent}
We designed a four-wheeled simulated robot Gazebo model. The robot's dimensions are $20\times20\times10$ cm and it is equipped with an event-based camera (see Section \nameref{sec:vision}) and the husky differential motor controller plugin. The \ac{BIBI}\cite{falotico2017NRP} connects the robot with the collision avoidance network implemented in NEST (see Section~\nameref{subsection:GFnetwork}). The connections consist of one transfer function from the vision sensor to the \ac{SPTC} population and another one from the \ac{MOT} population to the differential motor controller as well as two Poisson input spike sources. 
The first transfer function sends visual input events. 
The second transfer function controls the agent's insect-inspired movement pattern. During inactivity of the \ac{MOT} populations the robot drives purely translatory with a maximum speed of 2.5 a.u/s. The movement velocity changes anti-proportionally to the environment's obstacle density as explained in the section \nameref{subsubsec: cluttered}. When one of the two \ac{MOT} populations spikes the robot fixes its forward velocity to 0.38 a.u/s and turns either to the left or to the right with an angular velocity of 4 $^{\circ}/s$.
The two Poisson spike source populations send spikes with a medium spike rate of 100 Hz to the inverse, soft \ac{WTA} population and the \ac{ET} neuron (For more details see Table \ref{tab:connections} and Table \ref{tab:parameters}). 

\subsection{\ac{sEMD} characterisation}
\label{subsection:characterisation}

For the \ac{sEMD} characterisation we stimulated an event-based camera with a \SI{79}{\degree} lens (see Section~\nameref{sec:vision}) using square-wave gratings with a wavelength of \SI{20}{\degree} and various constant velocities (from 0.1 to 10 Hz). These recordings were performed in a controlled environment containing an event-based camera, an LED light ring and a moving screen which projects exchangeable stimuli (see Figure \ref{fig:Box}). The controllable light ring illuminates the screen. The camera's lens is positioned in the light ring's centre to ensure a homogeneous illumination of the pattern. The screen itself is moved by an Arduino controlled motor. During recordings, the box can be closed and thus be isolated from interfering light sources. The contrast refers to absolute grey-scale values printed on white paper to form the screen. However, given the printed contrast we calculated the Michelson contrast as follows:
\begin{equation}
\label{eq:Contrast}
     \frac{I_{max} - I_{min}}{I_{max} + I_{min}} = \frac{I_{max} - I_{max}(1-C_{printed})}{I_{max} + I_{max}(1-C_{printed})} = \frac{C_{printed}}{2-C_{printed}} 
\end{equation}

To show the model's robustness to a wide range of environments, we varied the following three parameters in the recordings: The illumination, the grating velocity and the grating's contrast (see Table \ref{tab:parametersSEMD}). Each possible parameter combination was recorded three times, with a recording duration of four seconds, to allow statistical evaluation of the results. The event-based camera was biased for slow velocities. 

The model (see Figure~\ref{fig:gap_network} the first three populations) was simulated in Nest with the connections and neuron parameters defined in Table \ref{tab:connections} and Table \ref{tab:parameters} respectively. The network was simulated for four seconds, receiving the events as emitted by the event-based camera as spike-source array input. To define a response to the various stimuli, from the simulation results, the mean population activity of the preferred direction and null direction population were calculated (see Figure \ref{fig:network_response}e). For the closest comparability to the biologically imposed environment parameters, we chose to compare and discuss the \ac{sEMD}'s velocity tuning curve for a grating contrast of \SI{100}{\percent} and an illumination of \SI{5000}{lux}.

\subsection{Closed-loop simulation in environments}
\label{subsection:environments}

\begin{figure}[h!]
\includegraphics[width = 0.6\textwidth]{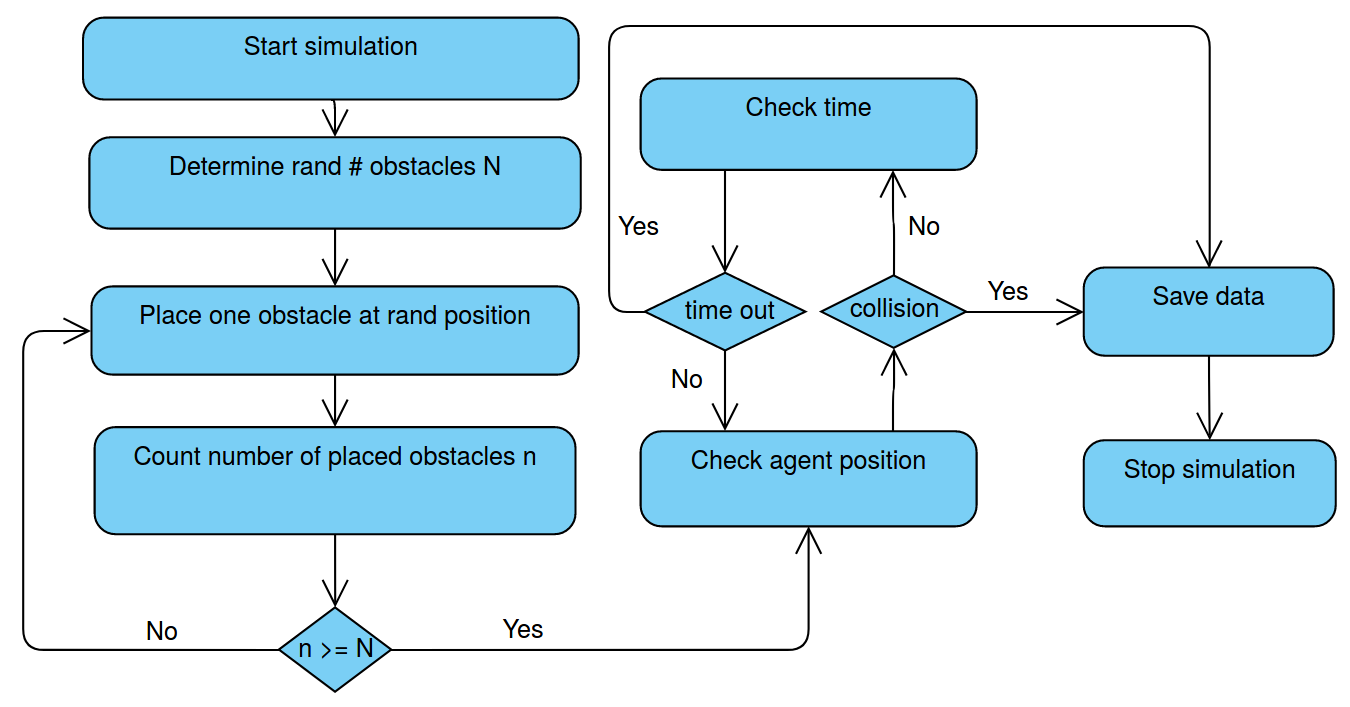}
\caption{State machine to create the cluttered environment and check the agent's collision avoidance performance.}
\label{fig:statemachine}
\end{figure}

Five different environments were designed to evaluate the agent's performance, a cluttered environment with randomly distributed obstacles sizing $1 \times 1$ m, an environment with two arenas connected by two gaps of variable size, a tunnel with varying width, an empty box environment and a narrowing tunnel. No obstacles were placed in a radius of two meters around the agent's start point so that the system can reach a stable state of activity before confronted with the first object. At obstacle densities higher than 35 percent the agent stays at its start point since no obstacle free direction can be detected anymore. Therefore, we limited the tested obstacle density range to 0 up to 38 percent. All obstacles placed in the environment including walls were covered with vertical black-and-white square-wave-gratings.
\\
A state-machine was written within the Neurorobotics Platform environment to automatise the experiments. The state-machine consists of eight states as shown in Figure \ref{fig:statemachine}

 Additionally, a virtual coach script was composed which starts and stops the single runs in a for-loop. After creating the simulation environment the virtual coach script starts the simulation for 10 seconds so that the state machine becomes activated. After that the simulation stops for five minutes which are long enough for the state machine to place all objects in the environment. When five minutes have passed the simulation is activated again and the agent starts moving through the environment. 
 CSV files containing the spiking data of the network, the robot position and angular alignment as well as the placement of the objects in the arena were saved for all experiments. 100 data points were collected for the collision avoidance experiment in a cluttered environment with adaptive velocity, 70 data points were collected for the experiment with fixed velocity (see Figure \ref{fig:behaviours}f,i \ref{fig:clearance},\ref{fig:velocity}  ). The tunnel centering experiment, gap entering experiment and all other simulation experiments in the appendix were repeated three times for each individual configuration (see Table \ref{tab:parameterssim}).
 \\
 Obstacle densities were calculated by plotting the cluttered environment and counting the number of pixels occupied by the objects.
 The occurrence of collisions was also measured visually by plotting the cluttered environment with the robot's trajectory while considering the agent size and angular alignment. Since the $can\textunderscore collide$ feature of the objects in the cluttered environment was turned off the agent moves through the obstacles when colliding. Therefore, an overlap of obstacle and robot can be interpreted as a collision.
 The collision avoidance run was marked as failed when such an overlap occurred and the first time of overlap was noted as collision time. Since there is no physical collision the robot's size can be varied during the analysis to evaluate the effect of agent size on the performance. To enhance the comparability of the robotic system to the biological role model, flying insects, we normalised all distance measures by dividing them by the chosen robot's size of 40x40 centimeters. The normalised distance measures were complemented with an arbitrary unit (a.u.).
 
\subsection{Data Availability Statement}

The data generated during this study will be available at dataverse.nl/dataset.xhtml?persistentId=doi:10.34894/QTOJJP.

\subsection{Code Availability Statement}

The code generated during this study will be made available at dataverse.nl/dataset.xhtml?persistentId=doi:10.34894/QTOJJP.

\section{Acknowledgements}
The authors would like to thank Daniel Gutierrez-Galan and Florian Hofmann for their technical support. The authors would also like to acknowledge the SpiNNaker Manchester team for their help with the sEMD implementation and the robot setup. Furthermore the authors acknowledge the Neurorobotics Platform team for their technical support.

\section{Author contribution}
T.S., E.C. and E.J. conceived and designed the experiments. M.B.M., T.S. and E.J. designed and optimised the tested algorithm. T.S. and E.J. carried out and analyzed the experiments. M.B.M., O.J.B., E.C. and M.E. developed the original sEMD model. O.J.B., M.B.M, T.S., E.J., E.C. and M.E. wrote the manuscript.

\section{Competing Interests}
The authors declare no competing interests.

\bibliography{}

\newpage
\appendix

\counterwithin{figure}{section} 
\section{Appendix}
\subsection{sEMD Characterization Setup}

\begin{figure}[h!]
\centering
\includegraphics[height = 0.2\textwidth]{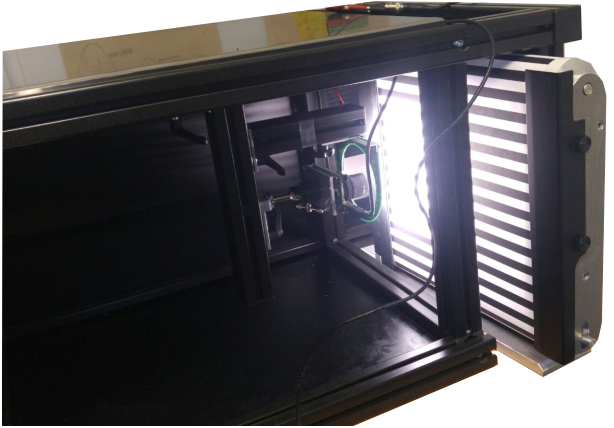}
\caption{Controlled environment for the recordings of the grating. The screen can move either from bottom to top or top to bottom. The upper roll of the screen contraption is driven by an Arduino controlled stepper-motor. The LED - ring illuminates the screen and the event driven camera is located in its center. }
\label{fig:Box}
\end{figure}

To ensure repeatability and reproducibility we recorded the grating in a controlled environment, see Figure \ref{fig:Box}. The \ac{DVS} is mounted in a light sealed box, with a variable distance to the screen. An LED-ring (with 32 LEDs) homogeneously illuminates the \ac{DVS}'s field of view. The LEDs them self are controlled by an external power-source. The moving screen consists of a thick paper tube, glued together at the ends with double-sided adhesive tape. This tube is clamped over two horizontally mounted cylinders. The lower cylinder is mounted with a floating bearing in the y-direction. The upper cylinder is driven by a stepper motor controlled by an Arduino Uno and translates its movement to the screen. The possible velocities of the screen range from \SI{23}{\milli\meter\per\second} to \SI{210}{\milli\meter\per\second}. The grating itself is printed on dull thick paper forming the paper-tube and stored in the dark to avoid fading out.

\subsection{\ac{sEMD} Implementation on SpiNNaker}

\begin{figure}[h!]
\begin{minipage}[]{0.48\textwidth}
\subfloat[ \label{}]{\includegraphics[width = \textwidth]{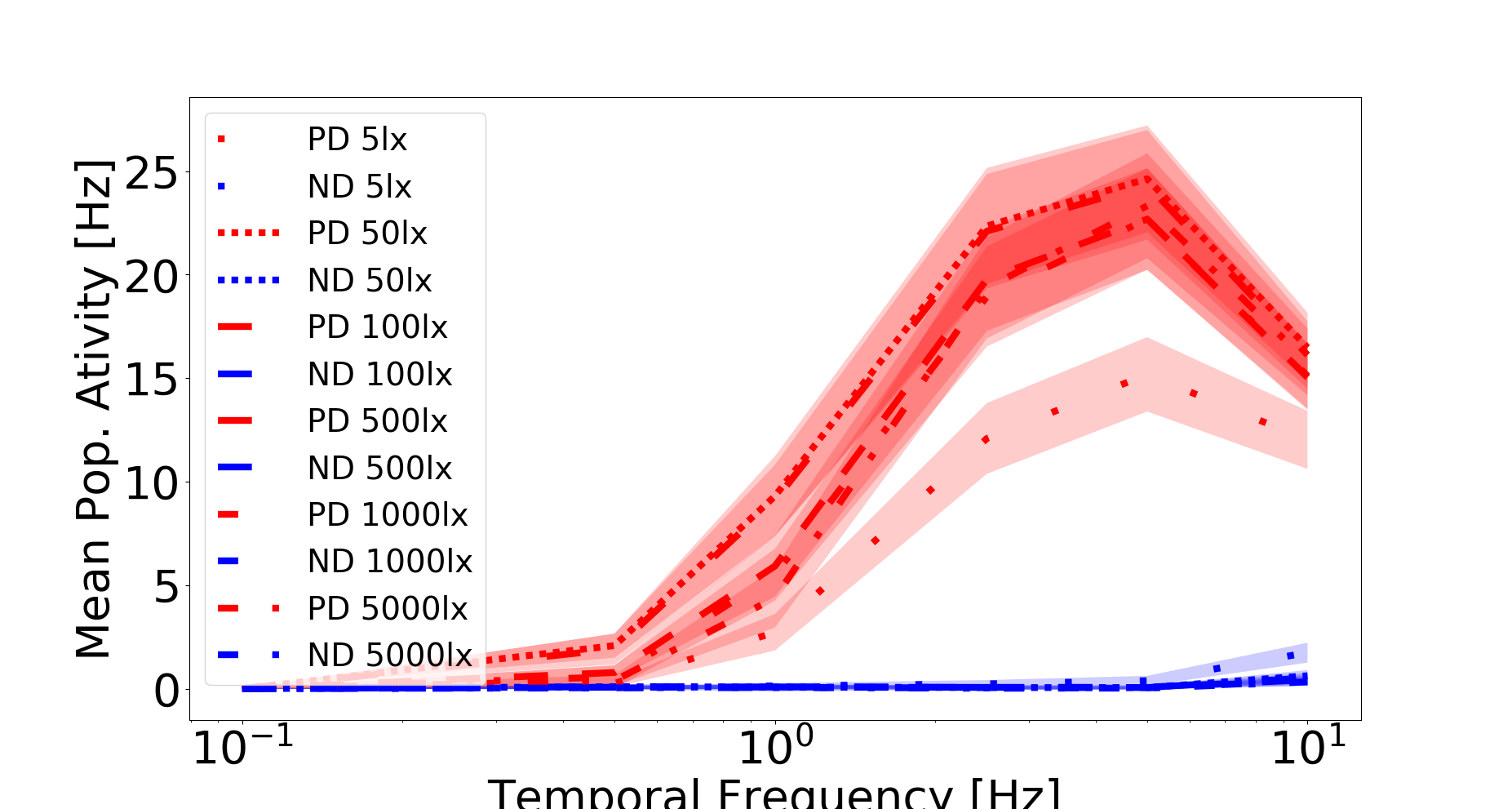}}
\end{minipage}
\begin{minipage}[]{0.48\textwidth}
\subfloat[ \label{}]{\includegraphics[width = \textwidth]{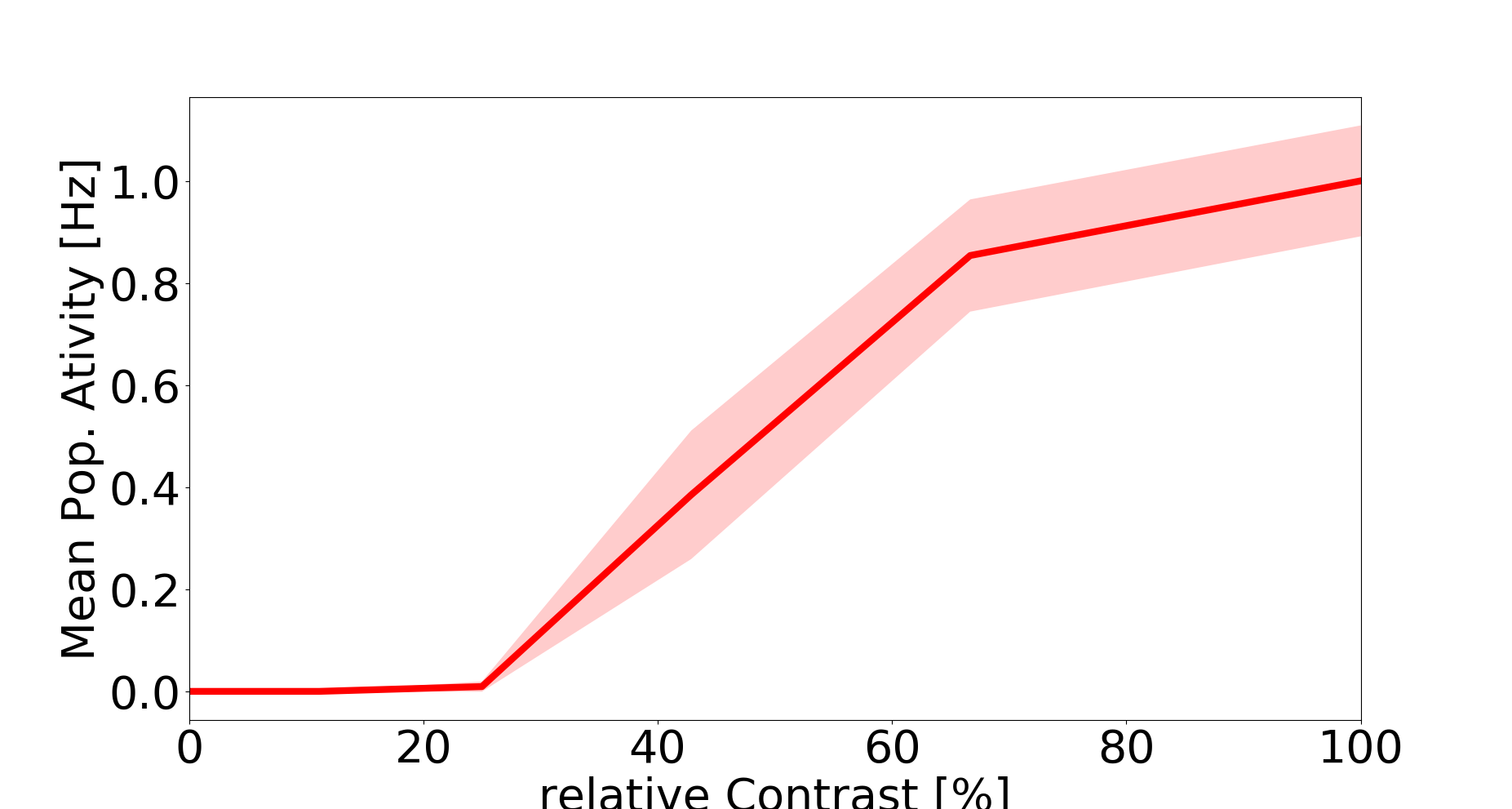}}
\end{minipage}
\caption{\ac{sEMD} population response on SpiNNaker for varying illuminations and contrasts. a) Normalised \ac{sEMD} population preferred direction and null direction response for \SI{100}{\percent} contrast and all illuminations from \SI{5}{lux} to \SI{5000}{lux}. b) Normalised preferred direction response for \SI{5000}{lux} illumination over contrasts varying from \SI{0}{\percent} to \SI{100}{\percent} at a temporal frequency of 5 Hz. For further information on the model parameters see Table \ref{tab:SpiNNparameters}.}
\label{fig:IlluminationAndContrast}
\end{figure}

To demonstrate the \ac{sEMD}'s wide range of operation and applicability on multiple platforms, we characterised the model's behaviour on SpiNNaker. We further investigated the \ac{sEMD}'s robustness regarding contrast and illumination. Figure \ref{fig:IlluminationAndContrast} a) shows that the model operates well in a wide range of illuminations at \SI{100}{\percent} contrast and produces similar velocity tuning curves on SpiNNaker and NEST (see Figure \ref{fig:network_response}e for comparison). Regarding the contrast sensitivity, we found that with the given parameter set, the model reaches half activity at a relative contrast of \SI{45.9}{\percent} (see Figure \ref{fig:IlluminationAndContrast} b) at \SI{5000}{lux} illumination and temporal frequency of 5 Hz. Thus the applicability of the model is limited by the occurring contrast but the offered range is still high and can possibly be improved by the implementation of contrast normalisation. 

\subsection{The Motion-Vision Network}
\label{sec:appendix_connectivity}
One very important parameter for collision avoidance is the knowledge of the own body size. Orchid bees with a wingspan of approximately 20 mm avoid to pass circular apertures smaller than 25 mm because of a too high collision risk. Some kind of self-representation in the bee's brain has to drive the insect's decision that the gap is too small for it. \cite{Baird2016gap}. Similarly, we can tune the connectivity of our \ac{SNN} to indirectly include relevant body size information. Our neural network model needs to consider its own body measures when moving through a gap. This decision process to move or not to move through a gap can be purely driven by the agent's relative perception of the gap. In our collision avoidance network this perception is modifiable by a change of the synaptic connections between the integrator neuron population and the inverse \ac{WTA} population. \ac{OF} is encoded in a retinotopical map of the integrator neuron population. This neuron population is initially one-to-one connected to the inverse \ac{WTA} network. By connecting the integrator neuron to its accordant inverse \ac{WTA} neuron and its closest neighbours the size of the perceived \ac{OF} caused by an object increases. Therefore, small gaps between objects are closed with increasing number of neighbouring \ac{INT} to inverse \ac{WTA} connections which leads to an increase of a perceived gap's minimum size.  The angle occupied by a gap has to be bigger than $gap_{min}$ to be considered a movement direction as shown in Equation \ref{eq:gap_angle}. $\alpha_{INT}$, the angle of perception of a single \ac{INT} neuron, amounts to \texttildelow\SI{2.2}{\degree} while $n_{connect}$ represents the number of neighbouring connections.  

\begin{equation}
\label{eq:gap_angle}
    gap_{min} = (2 \times n_{connect} + 1) \times \alpha_{sEMD}
\end{equation}

\begin{figure}[h!]
\begin{centering}
\includegraphics[width = 0.8\textwidth]{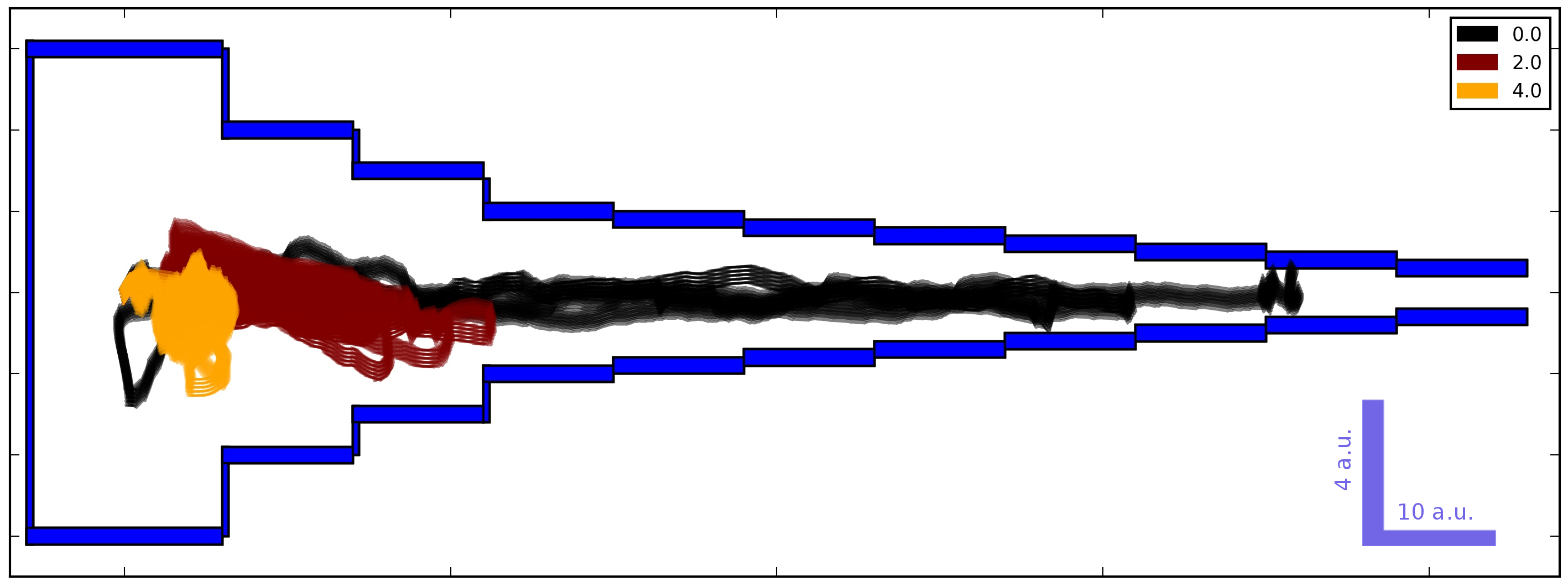}
\caption{Agent's trajectories in a narrowing corridor with varied connectivity between \ac{INT} and inverse \ac{WTA} population as explained in section \nameref{sec:appendix_connectivity}. Legend refers to number of neighbouring connections. Simulated robot's start point is on left side.}
\label{fig:corridor_1}
\end{centering}
\end{figure}

We evaluated how the minimum gap size entered by the robotic agent changes with the \ac{OF} perception. As expected, with increasing number of neighbouring connections small gaps were not entered anymore (see Figure \ref{fig:corridor_1}). By fixing the number of neighbouring connections to 4 nearest neighbours for all following experiments the robot wouldn't enter too small gaps but would still be able to navigate through larger corridors.

%Besides an overestimate of a gap's size the late perception of an obstacle can lead to an unintended collision because the agent's reaction time is too slow. Hence, an early detection of obstacles plays an important role in the context of collision avoidance. On the other side, the detection of only close-by objects allows the robot to take clear decisions not distracted by far-away objects. Hence, the right obstacle detection time range has to be found to enable a robust performance.In our collision avoidance model the detection time is notably influenced by the \ac{sEMD}'s facilitatory time constant $tau_{fac}$ (for detailed circuit-level response characterisation please refer to \cite{Milde2018sEMD}). The facilitatory time constant defines together with the input weight the minimum time-to-travel encoded by the \ac{sEMD}. By increasing $tau_{fac}$ the perceptual field extends towards weaker apparent motion stimuli. Therefore, the robotic agent detects far away objects earlier. We tested this hypothesis by varying the robot's $tau_{fac}$ in a small arena with square-wave grating on the walls. As expected, the robot's wall distance increased proportionally to $tau_{fac}$ because of an earlier detection time (see Figure \ref{fig:arena}a,c). We fixed $tau_{fac}$ to 10 milliseconds in all following experiments to enable an early detection of obstacles.

\begin{figure}[t!]
\begin{minipage}[]{0.48\textwidth}
\subfloat[ \label{}]{\includegraphics[width =0.9\textwidth]{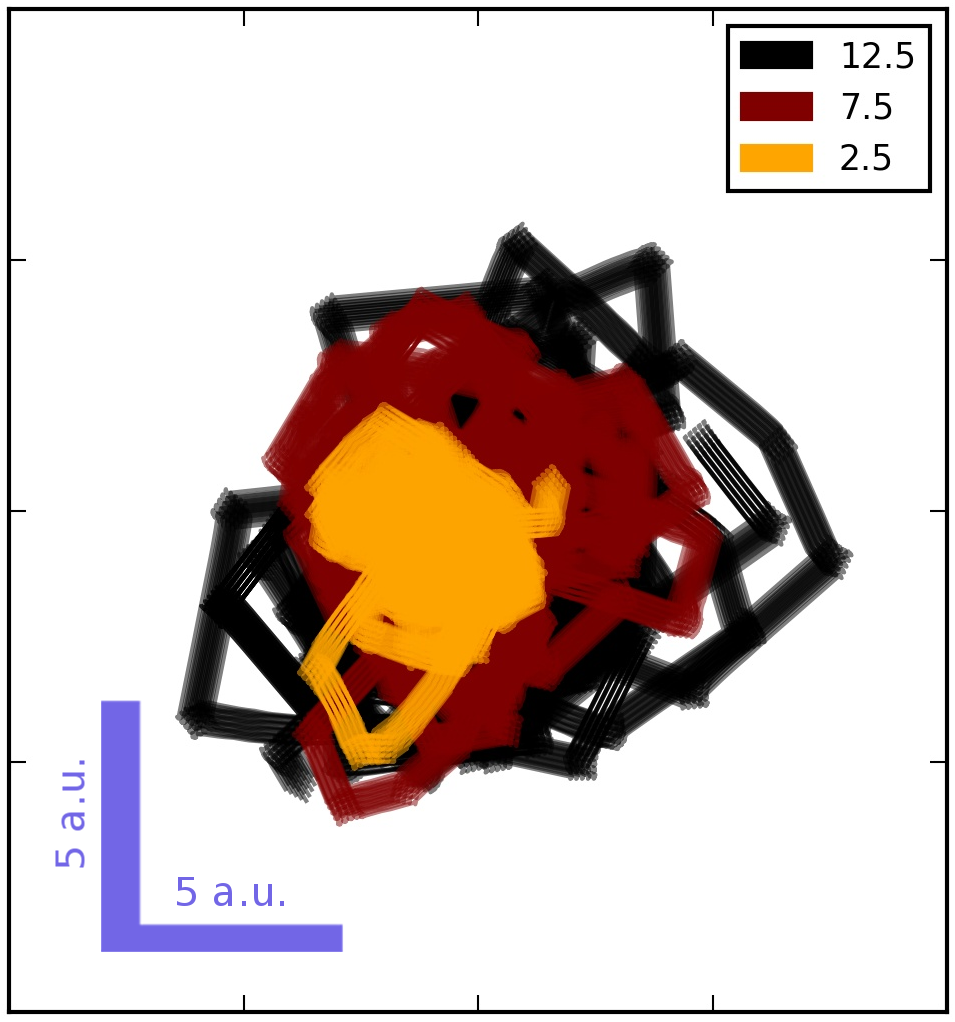}}
\end{minipage}
\begin{minipage}[]{0.48\textwidth}
\subfloat[ \label{}]{\includegraphics[width=0.9\textwidth]{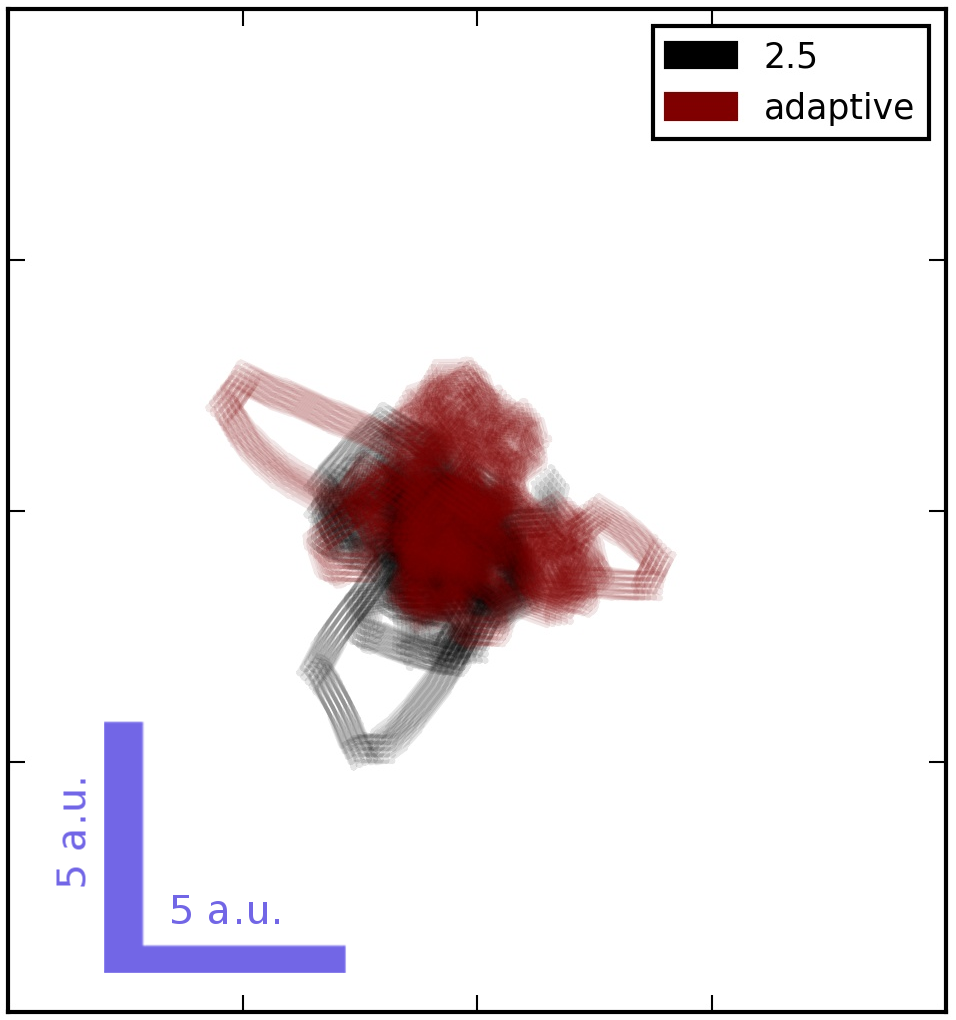}}
\end{minipage}
\caption{Agent's trajectories in an empty box. a) Agent's trajectory with different fixed intersaccadic velocities in a.u./s. b) Trajectories with parameters used for cluttered environment experiment in section \nameref{subsubsec: cluttered} with fixed and adaptive intersaccadic velocity in a.u./s.}
\label{fig:arena}
\end{figure}

\subsection{The Movement Behaviour}
\label{sec:appendix velocity}
When exposed to a densely cluttered environment, a narrow tunnel or a nearby object flying insects decrease their movement velocity. This mechanisms reduces the agent's collision probability by an increase in the time-of-flight. The agent has more time to react and turn away from the potential threat due to its lower speed. We tested the effect of a change in velocity with the agent in an empty arena. As expected, the simulated robot's minimum wall distance was increasing with lower velocities (see Figure \ref{fig:arena}a,b). Therefore, an adaptive obstacle density dependent velocity can be a helpful tool to increase the agents working range towards higher obstacle densities.

\subsection{Gap finding behaviour in cluttered environments}
\label{sec:appendix_clutter}
\begin{figure}[h!]
\includegraphics[width = 0.7\textwidth]{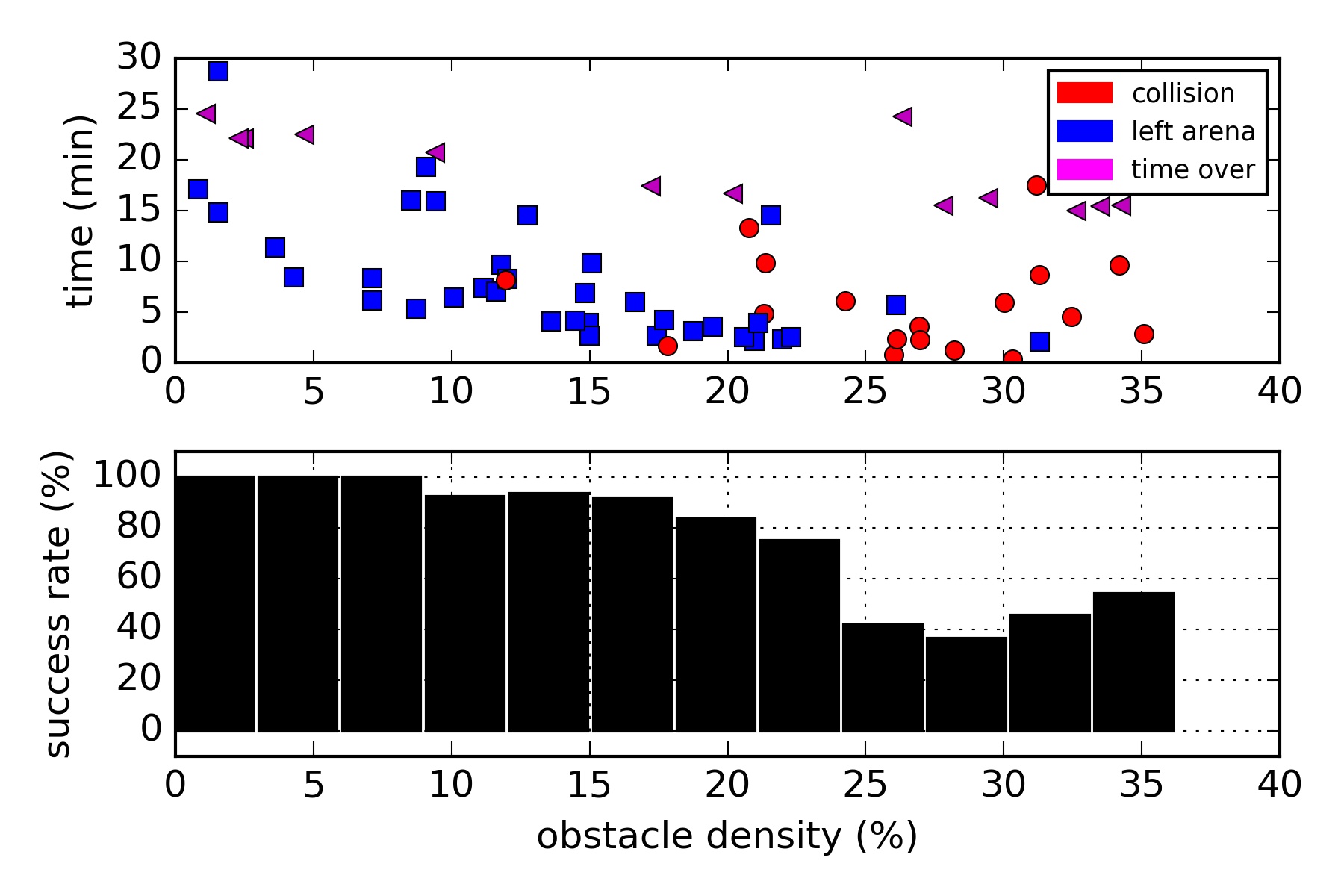}
\caption{Agent's behaviour in cluttered environments with the parameters from Table \ref{tab:parameters} and \ref{tab:connections} moving with a fixed intersaccadic velocity of 2.5 a.u./s. Top: Real world time at which the simulated robot leaves the arena, collides or the simulation time is over. Bottom: Agent's success rate, hence number of runs without collisions.}
\label{fig:cluttered}
\end{figure}

\begin{figure}[h!]
\includegraphics[width = 0.7\textwidth]{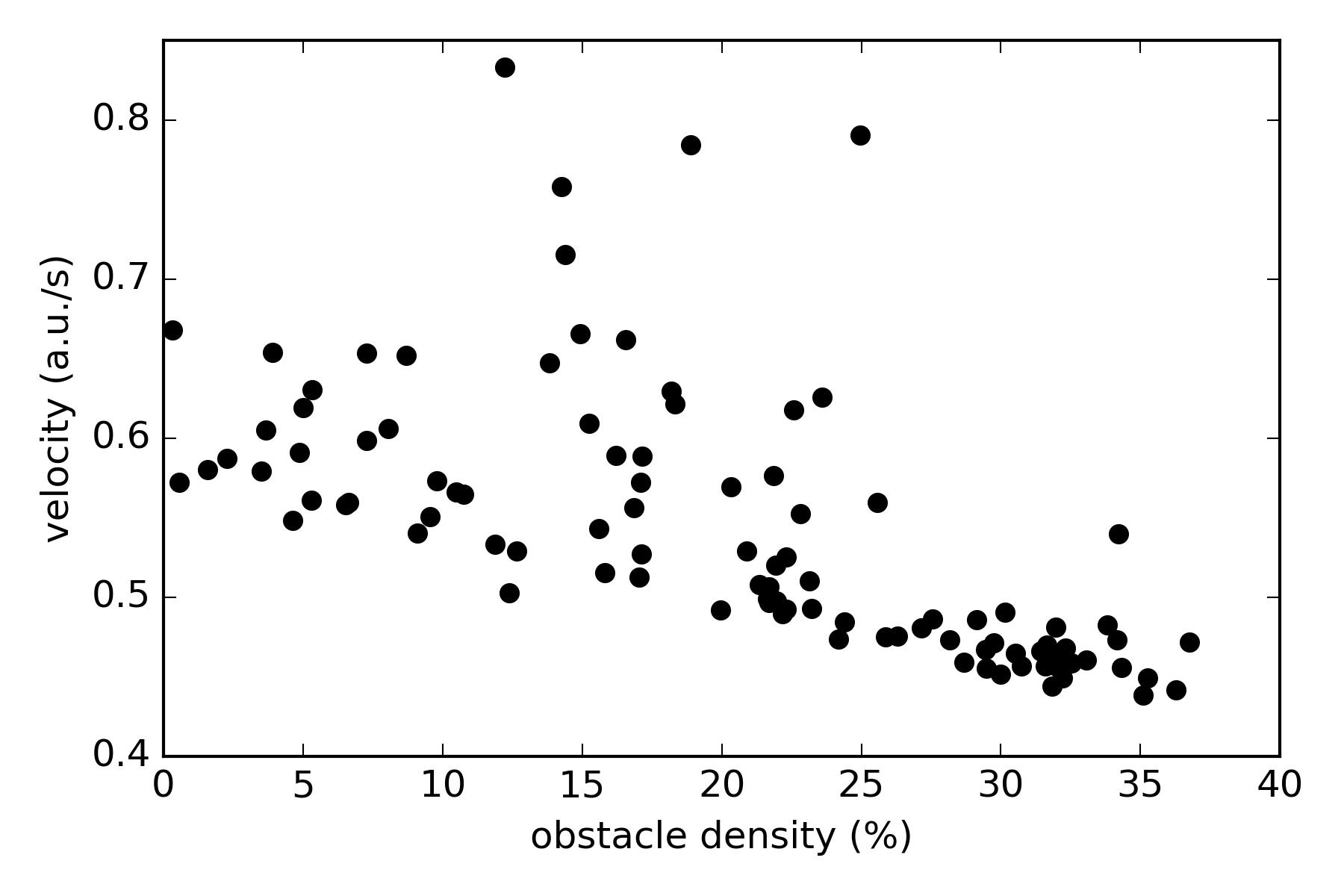}
\caption{Agent's mean velocity over obstacle density calculated for the data from Figure \ref{fig:behaviours}d}
\label{fig:velocity}
\end{figure}

Quantifying the relative motion perception and collision avoidance behaviour in controlled environments (see Figure~\ref{fig:Box} and \ref{fig:arena}) allows us to assess the fundamental capabilities of our agent. 
However, these tests do not fully capture conditions an agent will encounter in the real-world. 
These conditions include urban areas, indoors as well as outdoor forest environments.
A simple, yet effective test environment thus should be characterised with variable amount of clutter, i.e. obstacle density, of vertical obstacles placed in a random configuration.
\begin{figure}[b!]
\begin{minipage}[]{0.48\textwidth}
\subfloat[ \label{}]{\includegraphics[width = \textwidth]{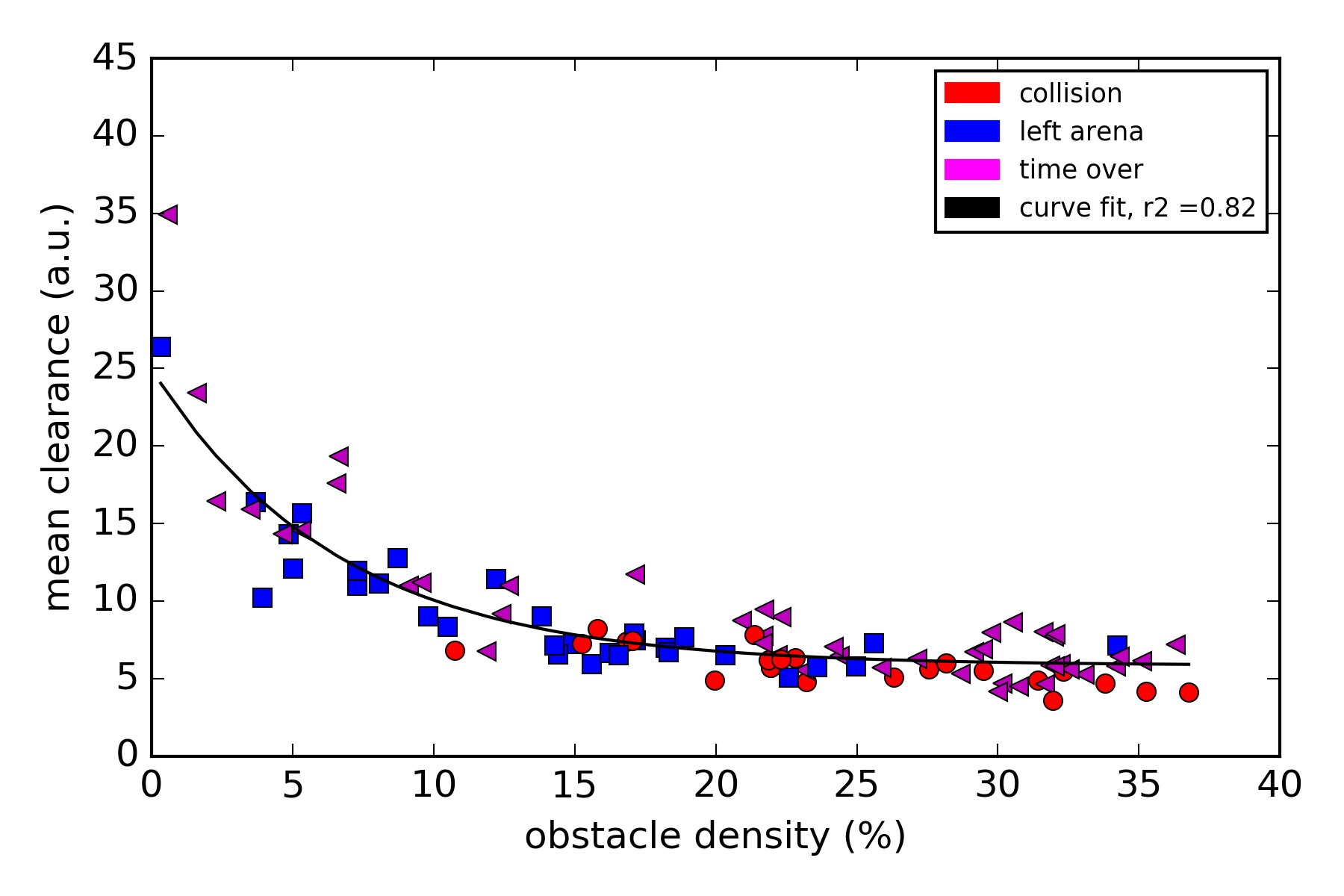}}
\end{minipage}
\begin{minipage}[]{0.48\textwidth}
\subfloat[ \label{}]{\includegraphics[width = \textwidth]{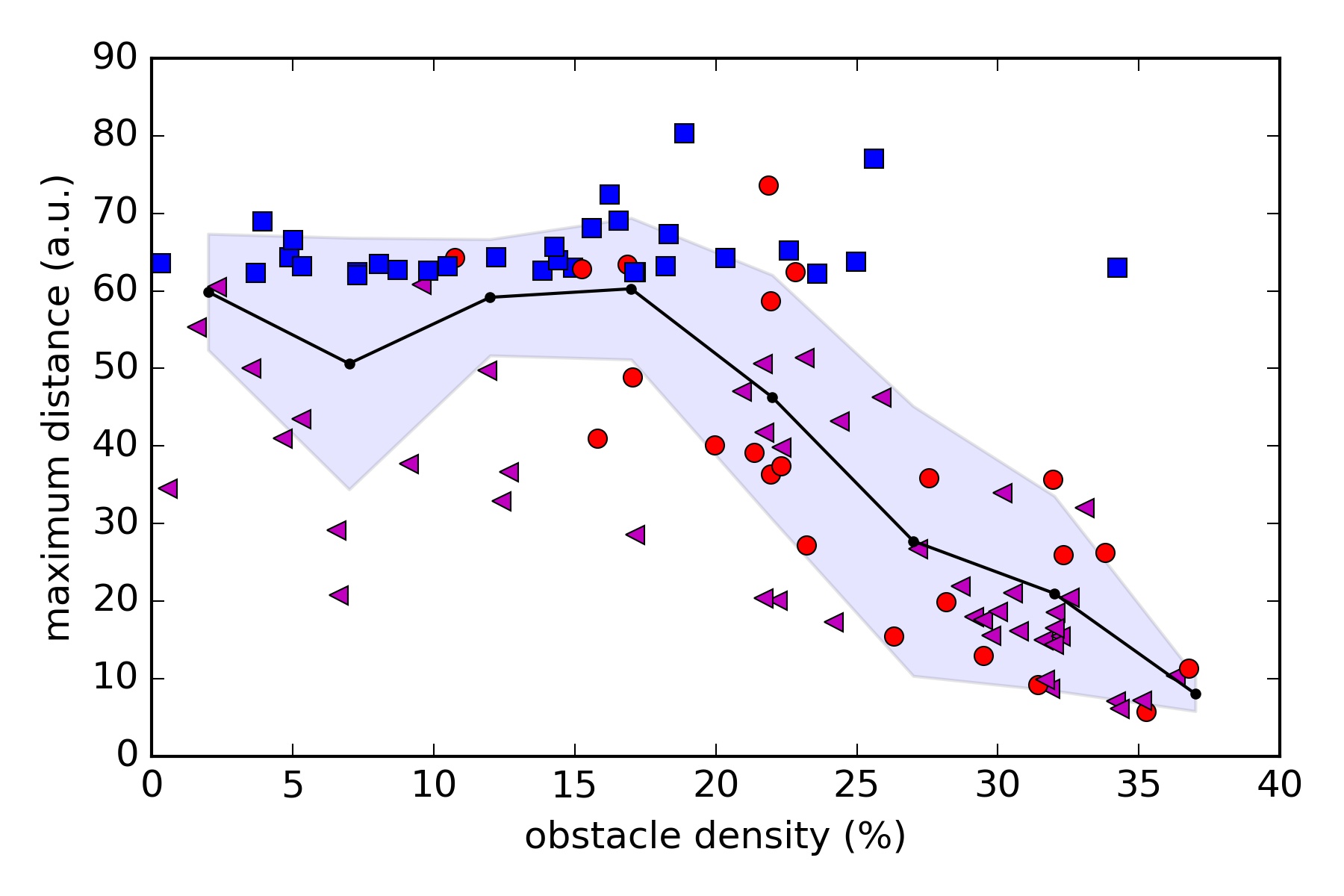}}
\end{minipage}
\caption{Agent's mean obstacle clearance and maximum distance to the start location calculated for the data from Figure \ref{fig:behaviours}d.}
\label{fig:clearance}
\end{figure}
We introduced the agent in an arena and varied the obstacle density from $0 \%$ up to $38 \%$ and measured the mean clearance (see Figure~\ref{fig:clearance} a) and the maximum distance (see Figure~\ref{fig:clearance} b) as a function of increasing obstacle density.
The mean clearance quickly drops from $25\ a.u.$ in roughly exponential fashion to a minimum of $5\ a.u.$.
If the obstacle density is greater than $\approx 15\ \%$, the mean clearance stays constant. 
However, the collision rate starts to increase (see Figure~\ref{fig:cluttered}).
Interestingly, due to the employed adaptive movement strategy the agent's velocity decreases almost linearly with increasing obstacle density (see Figure~\ref{fig:velocity}).
This adaptive behaviour ensures that despite high clutter the agent successfully identifies gaps in the environment and steers towards them and consequently avoids collisions with its surrounding.
\clearpage

\subsection{Corridor-centering in Real-World}

\begin{figure}[h!]
\begin{minipage}[]{0.48\textwidth}
\subfloat[ \label{}]{\includegraphics[width = 0.6\textwidth]{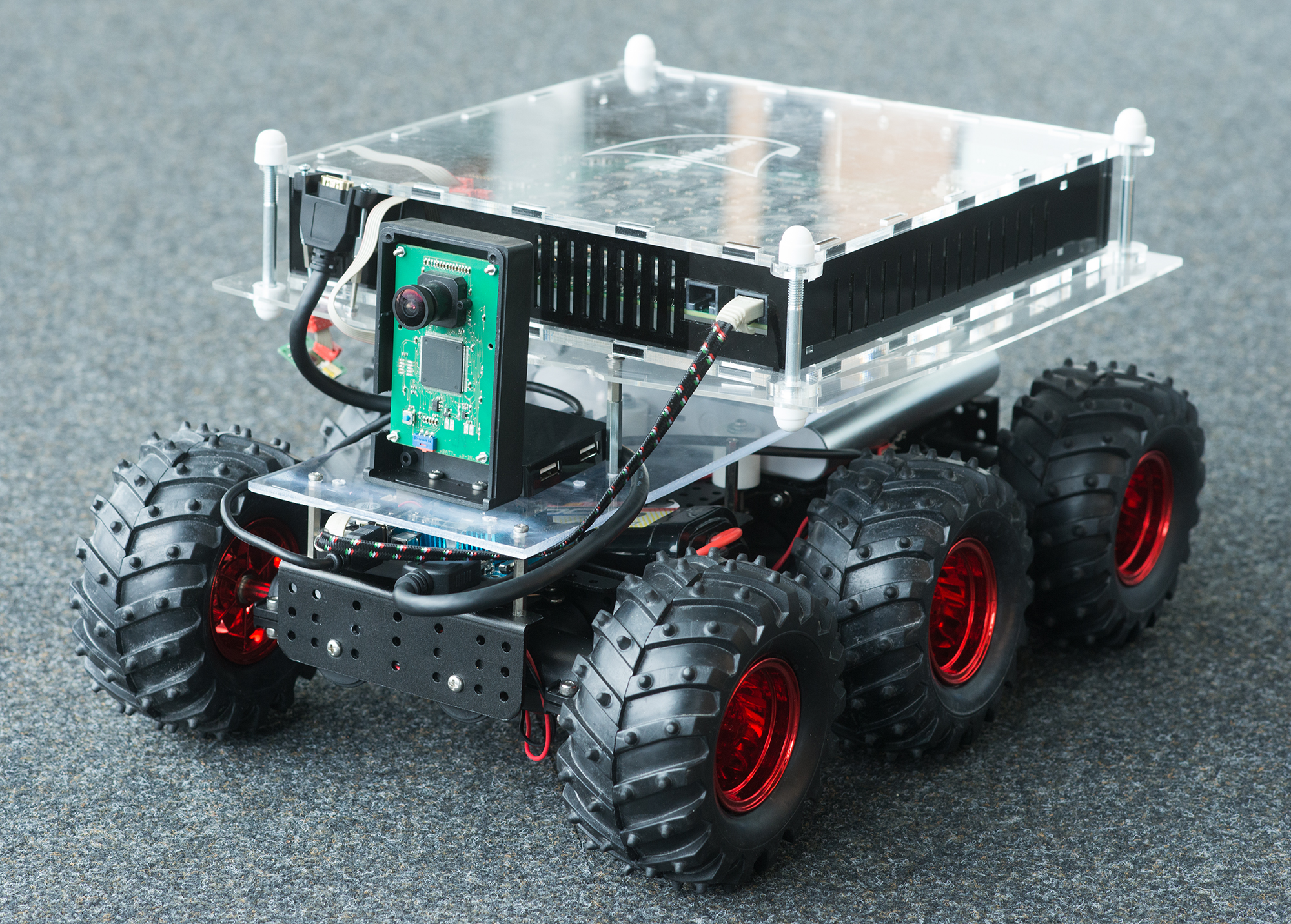}}
\end{minipage}
\begin{minipage}[]{0.48\textwidth}
\subfloat[ \label{}]{\includegraphics[width = 0.6\textwidth]{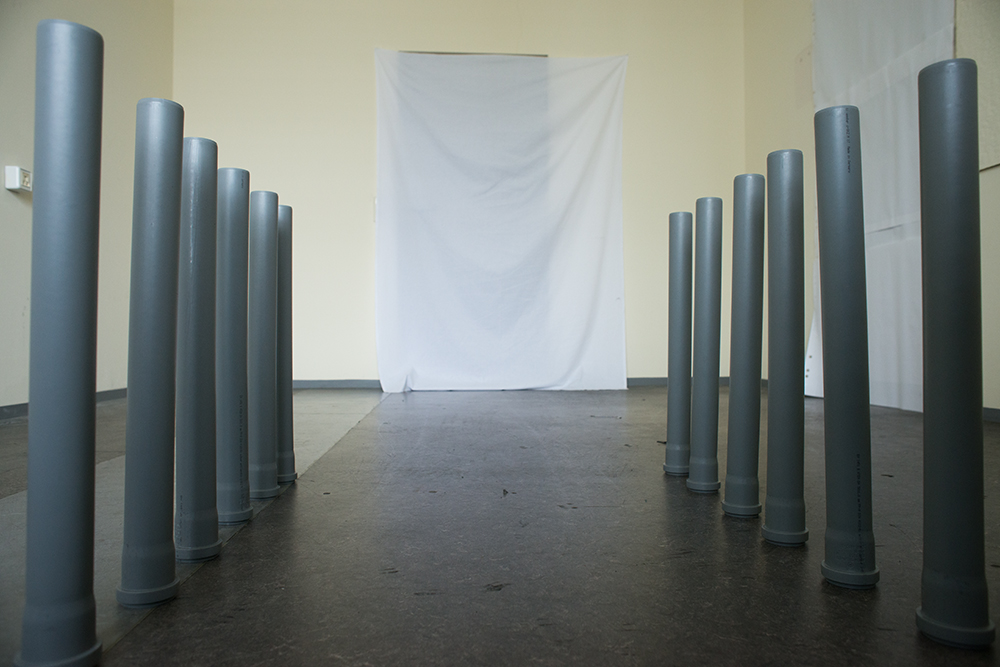}}
\end{minipage}
\caption{Robot and setup to conduct the real world experiment. a) The robot receives visual input from the embedded Dynamic Vision Sensor. The event-based camera sends its events to a SpiNN-5 board which simulates a simplified version of the collision avoidance network described in section \nameref{subsection:GFnetwork}. For more details on the real-world robotic implementation see section \nameref{subsec: real robot}. b) One meter long corridor.}
\label{fig:robot}
\end{figure}

\begin{figure}[h!]
\includegraphics[width = 0.75\textwidth]{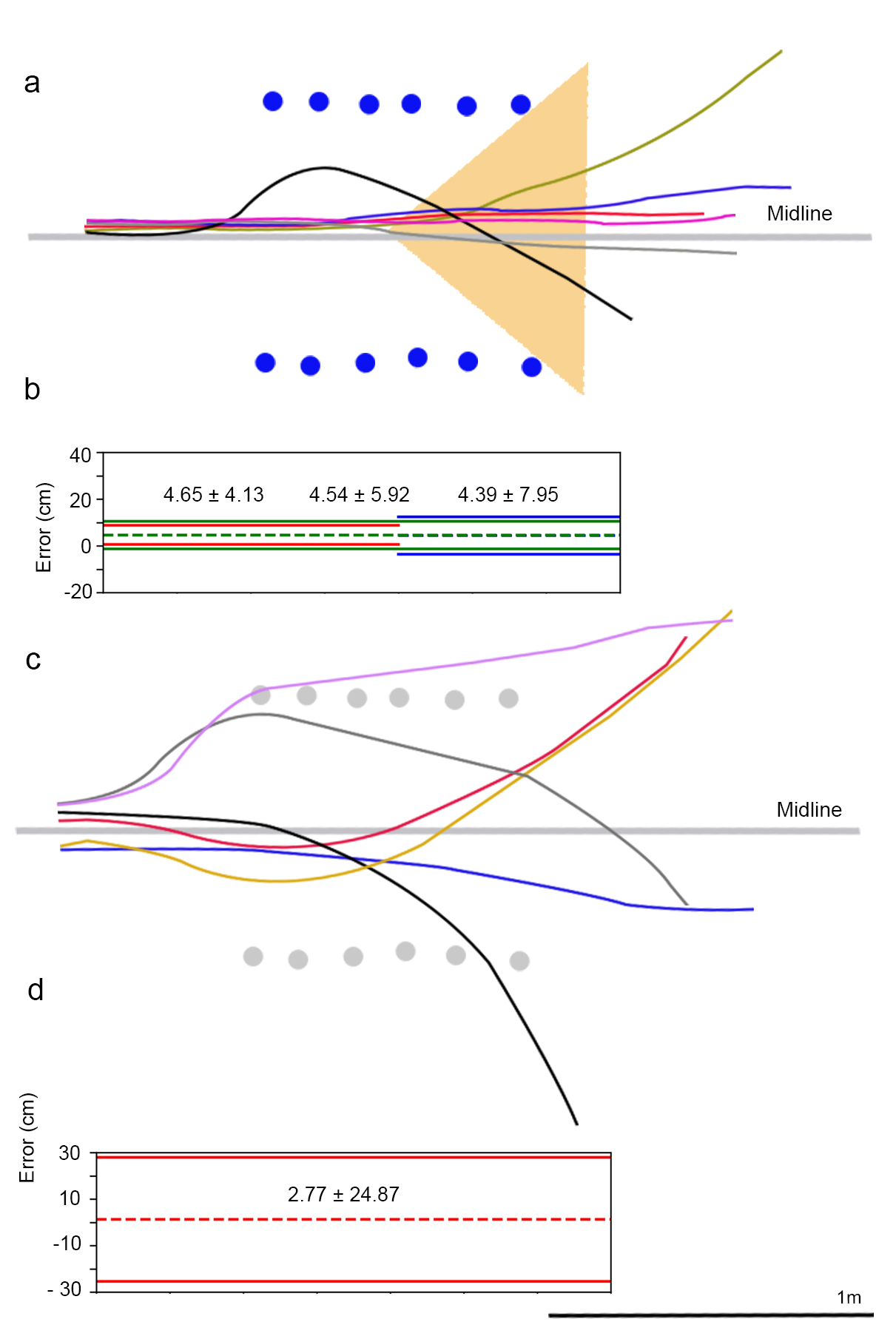}
\caption{Real world corridor centering experiment results. a) Robot's movement trajectories through corridor shown in figure \ref{fig:robot}. Movement direction is from left to right. Yellow triangle indicates position at which robot can not see the corridor anymore. b) Mean and standard-deviation for two regions from beginning until yellow triangle and from the yellow triangle until the end. The standard deviation increases for the second region since the robot does not see the corridor anymore. c) Control experiment of robot moving in environment without the corridor. d) Mean and standard-deviation of c).}
\label{fig:real_worl_results}
\end{figure}

To proof the real-time-capability and robustness of the \ac{SNN} on neuromorphic hardware, we evaluated the system in a real-world scenario. A robotic platform described in section \nameref{subsec: real robot} was assembled and tested in a corridor (see Figure \ref{fig:robot}). In five out of six cases the agent centers very well in the 80 centimeters wide corridor with a standard deviation from the corridor center between four and eight centimeters (see Figure \ref{fig:real_worl_results}a,b). In a control experiment with no corridor the robot was moving randomly into different directions showing that the robot's centering was caused by the corridor itself (see Figure \ref{fig:real_worl_results}c,d). 
%\end{linenumbers}

\newpage
\newpage
\section{Tables}
\begin{center}
\begin{table}[h!]
 \begin{tabular}{||c|c|c||} 
 \hline
 printed Contrast  & Temporal Frequency  & Illumination \\
  & (Hz) & (lux)\\
 [0.5ex] 
 \hline\hline
 0 & 0.1 & 5\\ 
 \hline
 0.2 & 0.5 & 50\\ 
 \hline
 0.4 & 1.0 & 100\\
 \hline
 0.6 & 2.5 & 500\\
 \hline
 0.8 & 5.0 & 1000\\
 \hline
 1.0 & 10.0 & 5000\\ [1ex] 
 \hline
 \end{tabular}
\caption{Parameters of grating recordings. Three four second recordings were made for each possible parameter-combination.}
\label{tab:parametersSEMD}
\end{table}
\end{center}
\begin{center}
\begin{table}[h!]
 \begin{tabular}{||c|c|c|c||} 
 \hline
 Simulation & Figures & Repetitions & Real time duration \\
  & & &(min)\\
 [0.5ex] 
 \hline\hline
 Clutter adaptive velocity & 2f,i, A.5, A.7a & 100 & 360\\ 
 \hline
 Clutter fixed velocity & 2i, A.6 & 70 & 360\\
 \hline
 Corridors & 2g,j,k,l & 3 per corridor width & 60\\
 \hline
 Real World Corridor & 2d & 6 & - \\
 \hline
 Gaps & 2e,h & 3 per gap size & 180\\
 \hline
 Narrowing Corridor & A.3 & 3 per configuration & 90 \\ 
  \hline
 Empty Box & A.4, A.7b & 3 per configuration & 30 \\[1ex] 
 \hline
 \end{tabular}
\caption{Parameters of simulations and real world experiment.}
\label{tab:parameterssim}
\end{table}
\end{center}
\newpage
\begin{center}
\begin{table}[h!]
 \begin{tabular}{||c| c c c c c c c c c c |c c||} 
 \hline
 Name & Type  & $C_{m}$ & $tau_{m}$ & $tau_{ref}$& $v_{reset}$ & $v_{rest}$ & $v_{thresh}$ & $tau_{syn\textunderscore E}$ & $tau_{syn \textunderscore I}$ & $I_{offset}$  & Popsize & $\#$Pop \\ &&&(nF) & (ms) & (ms) & (mV)  & (mV) & (mV) & (ms) & (ms) & (nA) & ($col \times row$) \\
 \hline\hline
 DVS & SSA & & & & & & & & & &$128\times 128$ & 1\\
 \hline
 SPTC & LIF & 0.25 & 20 & 1 & -85 & -60 & -50 & 20 & 20 & 0& $32 \times 32$ & 1 \\ 
 \hline
 sEMD & TDE & 0.25 & 20 & 1 & -85 & -60 & -50 & 20 & 20 & 0& $32 \times 32$ & 2\\
 \hline
 \hline
\end{tabular}

\begin{tabular}{||c|c| c c c c ||} 

    From & To & Weight (nA) & Connection type & Synapse type & delay (ms)\\ [0.5ex] 
    \hline\hline
    DVS & SPTC & 0.2 & (int(i/(128*4)*32) +int(i \SI{}{\percent} (128*4) / 3) to i) & excitatory & 1 \\ 
    \hline
    SPTC & TDE top-bottom & 0.2 & one\textunderscore to\textunderscore one & facilitator & 1 \\ 
    \hline
    SPTC & TDE top-bottom & 0.2 & i to i+32 & trigger & 1  \\
    \hline
    SPTC & TDE bottom-top & 0.2 & one\textunderscore to\textunderscore one & trigger & 1  \\
    \hline
    SPTC & TDE bottom-top & 0.2 & i+32 to i & facilitator & 0.1  \\
\hline
\end{tabular}
\caption{Neuron Parameters and Connections on SpiNNaker.}
\label{tab:SpiNNparameters}
\end{table}
\end{center}

\begin{center}
\begin{table}[h!]
 \begin{tabular}{||c| c c c c c c c c c c |c c||} 
 \hline
 Name & Type & $E_L$ & $C_{m}$ & $tau_{m}$ & $t_{ref}$& $tau_{syn\textunderscore exc}$ & $tau_{syn \textunderscore inh}$ &  $V_{th}$ & $V_{reset}$ & $V_{m}$ & Popsize & $\#Pop$\\ &&&(mV) & (pF) & (ms) & (ms) & (ms) & (ms) & (mV) & (mV) & (mV) & ($col \times row$)\\
 [0.5ex] 
 \hline\hline
 SPTC & LIF & -60.5 & 25 & 20 & 1 & 10 & 10 & -60 & -60.5 & -60.5 & $64\times 20$ & 1 \\ 
 \hline
 sEMD & TDE &-60.0 & 250 & 10 & 1 & 10 & 10 & -30 & -85 & -60 & $64\times 20$ & 2 \\
 \hline
INT & LIF & -70 & 250 & 20 & 1 & 5 & 5 & -40 & -70 & -65 &$ 64\times 1$ & 2 \\
 \hline
WTA & LIF & -65 & 250 & 20 & 1 & 5 & 80 & -50 & -68 & -65 & $64\times 1$ & 1 \\
 \hline
MOT & LIF & -65 & 250 & 20 & 2 & 5 & 5 & -50 & -68 & -65 & $96\times 1$ & 2\\
 \hline
GI & LIF & -65 & 250 & 30 & 2 & 40 & 5 & -50 & -68 & -65 &$ 1\times 1$ & 1\\
 \hline
OFI & LIF & -80 & 250 & 200 & 1 & 100 & 30 & -40 & -80 & -75 & $1\times 1$ & 1\\
\hline
ET & LIF & -65 & 250 & 20 & 1 & 5 & 80 & -50 & -68 & -65 &$ 1\times 1$ & 1\\[1ex] 
\hline
\hline
Name & Type & & Rate (Hz) & & & & & & & & Popsize & $\#Pop$  \\[0.5ex] 
\hline
\hline
POIS1 & Spike Source & & 100 & & & &  & &  &  &$ 64\times 1$ & 1\\
\hline
POIS2 & Spike  Source && 100 & & & &  & &  &  & $1\times 1$ & 1\\[1ex] 
\hline\hline
 \hline
\end{tabular}
\caption{Neuron Parameters from neurorobotics platform NEST network.}
\label{tab:parameters}
\end{table}
\end{center}
\begin{center}
\begin{table}[h!]
 \begin{tabular}{||c|c| c c c c ||} 
 \hline
 From & To & Weight (nA) & Connection type & Synapse type & delay (ms)\\ [0.5ex] 
 \hline\hline
 DVS NRP & SPTC & default & (i and i+1 and i+128 and i+ 129) to i & excitatory & 0.1 \\ 
 \hline
 DVS real world & SPTC & 0.002 & (i and i+1 and i+128 and i+ 129) to i  & excitatory & 0.1 \\ 
 \hline
 SPTC & TDE left-right & 4 & one\textunderscore to\textunderscore one & trigger & 0.1 \\ 
 \hline
 SPTC & TDE left-right & 4 & i to i+1 & facilitator & 0.1  \\
 \hline
 SPTC & TDE right-left & 4 & one\textunderscore to\textunderscore one & facilitator & 0.1  \\
 \hline
 SPTC & TDE right-left & 4 & i+1 to i & trigger & 0.1  \\
 \hline
 TDE right-left & INT right-left & 1 & i mod 64 to i & excitatory & 0.1 \\ 
 \hline
 TDE left-right & INT left-right & 1 & i mod 64 to i & excitatory & 0.1 \\ 
 \hline
 INT right-left & WTA & -5 & one\textunderscore to\textunderscore one & inhibitory & 0.1 \\ 
 \hline
  INT right-left & WTA & -3 & i to $i\pm$ 1 & inhibitory & 0.1 \\
 \hline
  INT right-left & WTA & -2 &
 i to $i\pm$ 2 & inhibitory & 0.1 \\
 \hline
  INT right-left & WTA & -1.5 &
 i to $i\pm$ 3 & inhibitory & 0.1 \\
 \hline
 INT right-left & OFI & $10^{-4}$ & $all\textunderscore to\textunderscore all$ & excitatory & 0.1 \\ 
 \hline
 INT left-right & WTA & -5 & one\textunderscore to\textunderscore one & inhibitory & 0.1 \\ 
 \hline
 INT left-right & WTA & -3 & i to $i \pm$ 1 & inhibitory & 0.1 \\ 
 \hline
 INT left-right & WTA & -2 & i to $i \pm$ 2 & inhibitory & 0.1 \\ 
 \hline
 INT left-right & WTA & -1.5 & i to $i \pm$ 3 & inhibitory & 0.1 \\
 \hline
 INT left-right & OFI & $10^{-4}$ & $all\textunderscore to\textunderscore all$ & excitatory & 0.1 \\ 
 \hline
 WTA(0-8) & MOT1 & 10 & i to 50 & excitatory & 0.1 \\ 
 \hline
 WTA(9-31) & MOT1 & 10 & i to 2i + 32 & excitatory & 0.1 \\ 
 \hline
 WTA(32-53) & MOT2 & 10 & 63 - i to 2i + 32 & excitatory & 0.1 \\ 
 \hline
 WTA(54-63) & MOT2 & 10 & i to 50 & excitatory & 0.1 \\ 
 \hline
 WTA & GI & 10 & all\textunderscore to\textunderscore all & excitatory & 0.1 \\ 
 \hline
 ET & MOT1 & 10 & 0 to 0 & excitatory & 0.1 \\ 
 \hline
 ET & GI & 10 & all\textunderscore to\textunderscore all & excitatory & 0.1 \\ 
 \hline
 GI & ET & -10 & all\textunderscore to\textunderscore all & inhibitory & 0.1 \\ 
 \hline
 GI & WTA & -10 & all\textunderscore to\textunderscore all & inhibitory & 0.1 \\ 
 \hline
 MOT1 & WTA & -30 & all\textunderscore to\textunderscore all & inhibitory & 0.1 \\ 
 \hline
 MOT1 & ET & -30 & all\textunderscore to\textunderscore all & inhibitory & 0.1 \\ 
 \hline
 MOT1 & MOT2 & -10 & all\textunderscore to\textunderscore all & inhibitory & 0.1 \\ 
 \hline
 MOT1 & Sensors & -30 & all\textunderscore to\textunderscore all & inhibitory & 0.1 \\ 
 \hline
 MOT1 & MOT1 & 10 & i to i + 1 & excitatory & 10 \\ 
 \hline
 MOT1 & MOT1 & -10 & one\textunderscore to\textunderscore one & inhibitory & 0.1 \\ 
 \hline
 MOT2 & WTA & -30 & all\textunderscore to\textunderscore all & inhibitory & 0.1 \\ 
 \hline
 MOT2 & ET & -30 & all\textunderscore to\textunderscore all & inhibitory & 0.1 \\ 
 \hline
 MOT2 & MOT1 & -10 & all\textunderscore to\textunderscore all & inhibitory & 0.1 \\ 
 \hline
 MOT2 & Sensors & -30 & all\textunderscore to\textunderscore all & inhibitory & 0.1 \\ 
 \hline
 MOT2 & MOT2 & 10 & i to i + 1 & excitatory & 10 \\ 
 \hline
 MOT2 & MOT2 & -10 & one\textunderscore to\textunderscore one & inhibitory & 0.1 \\ 
 \hline
 POIS1 & WTA & 1 & one\textunderscore to\textunderscore one & excitatory & 0.1 \\ 
  \hline
 POIS2 & ET & 0.3 & one\textunderscore to\textunderscore one & excitatory & 0.1 \\ [1ex] 
 \hline
\end{tabular}
\caption{Neuron connections from NEST network used in the neurorobotics platform. Note: There might be slight differences in the connection scheme when comparing Figure \ref{fig:gap_network} with this table. This is because Figure \ref{fig:gap_network} only serves for demonstration purposes. Always use the connections from this table to rebuild the network.}
\label{tab:connections}
\end{table}
\end{center}

\end{document}